\documentclass[runningheads]{llncs}
\usepackage[T1]{fontenc}
\usepackage{graphicx}
\usepackage{booktabs}
\usepackage[misc]{ifsym}

\usepackage{amssymb}
\usepackage{amsmath}
\usepackage[compatibility=false]{caption}
\usepackage{subcaption}
\usepackage{float}
\usepackage{lipsum}
\usepackage{algorithm}
\usepackage{algpseudocode}

\begin{document}

\title{ICLAD: In-Context Learning for Unified Tabular Anomaly Detection Across Supervision Regimes}

\titlerunning{ICLAD}

\author{Jack Yi Wei\inst{1,2} \and
Narges Armanfard \inst{1,2}}


\institute{Department of Electrical and Computer Engineering, McGill University, Canada 
\and
Mila - Quebec Artificial Intelligence Institute, Montreal, Canada
\\ \email{yi.wei4@mail.mcgill.ca, narges.armanfard@mcgill.ca}
}

\maketitle              

\begin{abstract}
Anomaly detection on tabular data is commonly studied under three supervision regimes, including one-class settings that assume access to anomaly-free training samples, fully unsupervised settings with unlabeled and potentially contaminated training data, and semi-supervised settings with limited anomaly labels. Existing deep learning approaches typically train dataset-specific models under the assumption of a single supervision regime, which limits their ability to leverage shared structures across anomaly detection tasks and to adapt to different supervision levels. We propose ICLAD, an in-context learning foundation model for tabular anomaly detection that generalizes across both datasets and supervision regimes. ICLAD is trained via meta-learning on synthetic tabular anomaly detection tasks, and at inference time, the model assigns anomaly scores by conditioning on the training set without updating model weights. Comprehensive experiments on 57 tabular datasets from ADBench show that our method achieves state-of-the-art performance across three supervision regimes, establishing a unified framework for tabular anomaly detection.

\end{abstract}

\section{Introduction}

Most real-world datasets are represented as tables, with rows corresponding to samples and columns representing attributes. Tabular anomaly detection aims to identify unusual rows within such datasets and has widespread applications in cybersecurity, healthcare, and industrial systems, where anomalies may indicate network intrusions~\cite{alrawashdehOnlineAnomalyIntrusion2016}, medical risks~\cite{yousefExploratoryRiskPrediction2024}, or equipment failures~\cite{ringlerUnsupervisedAnomalyDetection2025}.

Despite the recent popularity of deep learning, effective anomaly detection on tabular data remains challenging. Tabular data often consist of heterogeneous features with mixed continuous and discrete variables, and lack strong relation structures such as spatial, sequential or graphical dependencies~\cite{borisovDeepNeuralNetworks2022}. In tabular supervised learning, deep learning models have historically struggled to consistently outperform gradient-boosted decision trees~\cite{shwartz-zivTabularDataDeep2022,grinsztajnWhyTreebasedModels2022b,mcelfreshWhenNeuralNets2023b}. A similar trend holds for tabular anomaly detection, where numerous deep methods have been proposed but classical density-based methods remain competitive on standard benchmarks~\cite{hanADBenchAnomalyDetection2022a,livernocheDiffusionModelingAnomaly2023b}. These observations suggest that generic deep learning approaches often struggle to capture inductive biases well suited to tabular data, motivating approaches that can acquire tabular-specific inductive biases.

Beyond the challenge of tabular deep learning, tabular anomaly detection is further complicated by a diverse but fragmented set of supervision settings~\cite{pangDeepLearningAnomaly2021}. Recent deep learning methods typically assume the one-class setting, in which the training data is clean with all normal samples~\cite{bergmanClassificationBasedAnomalyDetection2019,xuFascinatingSupervisorySignals2023a,shenkarAnomalyDetectionTabular2021a}. However, in practice, datasets can contain an unknown proportion of unlabeled anomalies~\cite{chalapathyDeepLearningAnomaly2019}, and in many cases a number of anomalies can be labeled with little cost~\cite{ruffDeepSemiSupervisedAnomaly2019a}. Although some works treat the one-class scenario as a special case of semi-supervised learning~\cite{livernocheDiffusionModelingAnomaly2023b,ruffUnifyingReviewDeep2021b}, we distinguish three supervision regimes for clarity. We refer to the case of a clean training set as the one-class setting, the case of unlabeled but potentially contaminated training data as the unsupervised setting, and to that with a small number of labeled anomalies as the semi-supervised setting, where unlabeled samples are not assumed to be normal. While specialized methods exist for each regime~\cite{ruffDeepSemiSupervisedAnomaly2019a,qiuLatentOutlierExposure2022}, their objectives are typically tied to the assumed supervision setting, and models trained under one regime may not apply to others or fail to generalize robustly when applied across settings~\cite{livernocheDiffusionModelingAnomaly2023b}.

These limitations motivate an approach that can acquire inductive biases aligned to tabular anomaly detection and adapt inference behavior to different supervision regimes. This can be achieved through in-context learning, a paradigm that allows a model to adapt inference from the training data provided as a context, without updating model parameters. In the case of anomaly detection, a model can infer the supervision regime from the labels in the context and modulate the decision function accordingly. Such in-context adaptation can be enabled through meta-learning over a distribution of tasks. With an appropriate distribution of tabular tasks, meta-learning can impart inductive biases suitable for tabular data. Recent tabular foundation models~\cite{hollmannTabPFNTransformerThat2022a,hollmannAccuratePredictionsSmall2025b,quTabICLTabularFoundation2025a} provide evidence for this approach, demonstrating that meta-learning on synthetic data can capture shared structure across tabular classification and regression problems. Contemporaneous work~\cite{shenFoMo0DFoundationModel2025} explores in-context learning for anomaly detection, yet the ability of a single model to adapt to different supervision regimes in anomaly detection remains underexplored.

We propose In-Context Learning for Anomaly Detection (ICLAD), a foundation model designed to generalize across tabular datasets and supervision regimes. A key component of our approach is the construction of a distribution of synthetic tabular anomaly detection tasks that vary in dataset characteristics, supervision assumptions, and anomaly contamination levels. ICLAD is implemented as a transformer trained on this task distribution. After this single training phase, the model can be reused on new datasets without updating its parameters. At inference time, ICLAD conditions on the training data and their associated labels as context, enabling adaptation to new datasets and supervision regimes through in-context learning. In this way, ICLAD captures generalizable patterns across tabular anomaly detection tasks and provides a unified framework for one-class, unsupervised, and semi-supervised settings.

The main contributions of this work are summarized as follows:
\begin{enumerate}
    \item We introduce ICLAD, an in-context learning foundation model for tabular anomaly detection that performs task-level inference and generalizes across datasets and supervision regimes without retraining. To the best of our knowledge this work is the first to introduce a unified in-context anomaly detection model that generalize across anomaly detection supervision regimes.
    \item We show that this capability arises from meta-learning on a diverse distribution of synthetic tabular tasks spanning dataset properties, supervision, and contamination levels.
    \item We conduct an extensive empirical evaluation on ADBench~\cite{hanADBenchAnomalyDetection2022a}, a benchmark of 57 real-world tabular datasets, comparing ICLAD with a wide range of classical and deep anomaly detection methods across unsupervised, one-class, and semi-supervised settings.
\end{enumerate}

\begin{figure*}[t]
    \centering

    \begin{subfigure}[t]{0.32\textwidth}
        \centering
        \includegraphics[width=\linewidth]{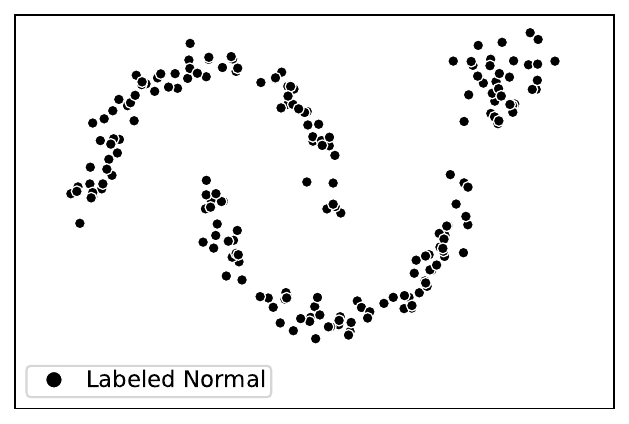}
        \includegraphics[width=\linewidth]{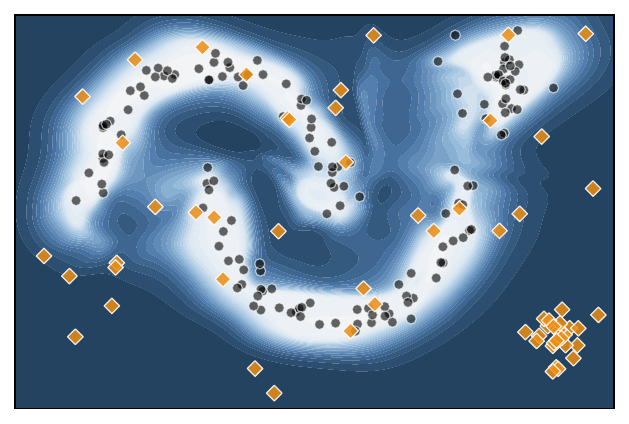}
        \caption{One-class setting}
    \end{subfigure}
    \begin{subfigure}[t]{0.32\textwidth}
        \centering
        \includegraphics[width=\linewidth]{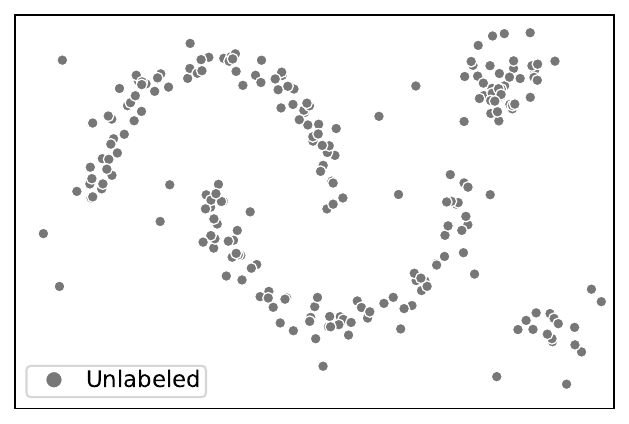}
        \includegraphics[width=\linewidth]{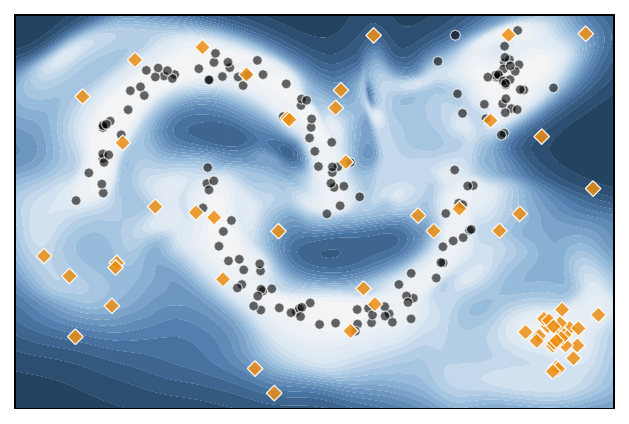}
        \caption{Unsupervised setting}
    \end{subfigure}
    \begin{subfigure}[t]{0.32\textwidth}
        \centering
        \includegraphics[width=\linewidth]{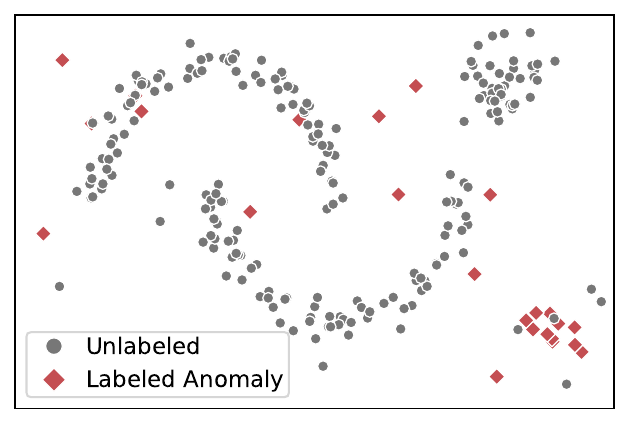}
        \includegraphics[width=\linewidth]{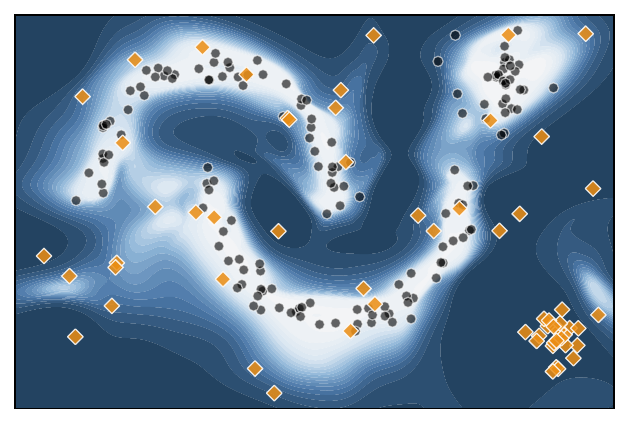}
        \caption{Semi-supervised setting}
    \end{subfigure}

    \caption{
    Each column corresponds to a training regime. The top row shows the training dataset, and the bottom row shows the test set and the anomaly-score landscape from ICLAD after conditioning on training dataset. In the test set, orange diamonds denote ground-truth anomalies and dark circles denote ground-truth normal samples. Darker regions indicate higher anomaly scores.
    }
    \label{fig:iclad_teaser}
\end{figure*}

\section{Related Work}

\subsection{One-Class and Unsupervised Tabular Anomaly Detection}

Classical approaches to tabular anomaly detection typically operate directly in the input space or through shallow transformations. These include distance- and density-based methods such as kNN~\cite{ramaswamyEfficientAlgorithmsMining2000}, LOF~\cite{breunigLOFIdentifyingDensitybased2000}, CBLOF~\cite{heDiscoveringClusterbasedLocal2003}, and Isolation Forest~\cite{liuIsolationForest2008}; one-class classification approaches such as OCSVM~\cite{scholkopfSupportVectorMethod1999}; subspace-based methods such as PCA~\cite{shyuNovelAnomalyDetection2003}; and statistical techniques such as HBOS~\cite{goldsteinHistogrambasedOutlierScore2012} and ECOD~\cite{liECODUnsupervisedOutlier2023}. Motivated by the success of deep learning, more recent work has explored using neural networks for tabular anomaly detection, including autoencoders~\cite{hintonReducingDimensionalityData2006}, Deep SVDD~\cite{ruffDeepOneClassClassification2018a}, LUNAR~\cite{goodgeLUNARUnifyingLocal2022b}, and various self-supervised representation learning approaches such as GOAD~\cite{bergmanClassificationBasedAnomalyDetection2019}, ICL~\cite{shenkarAnomalyDetectionTabular2021a}, MCM~\cite{yinMCMMaskedCell2023b}, NeutralAD~\cite{qiuNeuralTransformationLearning2021}, DRL~\cite{yeDRLDecomposedRepresentation2024a}, and SLAD~\cite{xuFascinatingSupervisorySignals2023a}. DTE~\cite{livernocheDiffusionModelingAnomaly2023b} introduces diffusion time as a proxy for outlier scoring, which is related to $k$-NN density estimation but with efficient neural approximations. While deep models offer greater representational flexibility, empirical studies suggest that they do not consistently outperform strong classical baselines on tabular datasets~\cite{livernocheDiffusionModelingAnomaly2023b,boumanUnsupervisedAnomalyDetection2024,hanADBenchAnomalyDetection2022a}. 

Another line of work in unsupervised tabular anomaly detection involves contamination-robust approaches. Most of the aforementioned methods, particularly deep methods, assume virtually clean training data aligned with the one-class setting. In the common scenario where there are unknown levels of contamination, these models often face deteriorating performance~\cite{livernocheDiffusionModelingAnomaly2023b}, motivating robust approaches such as the Minimum Covariance Determinant (MCD)~\cite{rousseeuwLeastMedianSquares1984} and the recent iterative refinement methods like Latent Outlier Exposure (LOE)~\cite{qiuLatentOutlierExposure2022}.

\subsection{Semi-Supervised Anomaly Detection on Tabular Data}

In the unsupervised setting, labeled anomalies can guide anomaly detection by providing prior information. In this case, anomaly detection can be framed as a semi-supervised learning problem related to positive-unlabeled (PU) learning. Representative methods include DevNet~\cite{pangDeepAnomalyDetection2019}, PreNet~\cite{pangDeepWeaklysupervisedAnomaly2023}, FEAWAD~\cite{zhouFeatureEncodingAutoencoders2022}, and GANomaly~\cite{akcayGANomalySemisupervisedAnomaly2019}. These methods rely on objectives that require labeled anomalies during training and are therefore tied to the semi-supervision regime. DeepSAD~\cite{ruffDeepSemiSupervisedAnomaly2019a} partially addresses this by extending Deep SVDD with the ability to utilize supervision, but still requires retraining when supervision levels change.

\subsection{Prior-data Fitted Networks}

Prior-data fitted networks (PFNs)~\cite{mullerTransformersCanBayesian2021} are models trained to perform in-context learning by meta-learning over a distribution of related tasks, often instantiated through synthetic datasets. PFNs amortize Bayesian inference over data-generating mechanisms into a single training stage, enabling in-context adaptation on new datasets without parameter updates. The representative PFN-based models for tabular data are TabPFN~\cite{hollmannTabPFNTransformerThat2022a,hollmannAccuratePredictionsSmall2025b} and TabICL~\cite{quTabICLTabularFoundation2025a}, which achieve strong performance on tabular classification and regression. These models demonstrate that in-context learning can capture shared structure across tabular tasks and generalize effectively to unseen datasets without retraining. Recently, a similar work Fomo-0D~\cite{shenFoMo0DFoundationModel2025} investigated a PFN framework to one-class anomaly detection, showing that outliers can be identified through in-context learning. However, a crucial direction remains under-explored, i.e., the capability of in-context learning model to adapt its inductive bias to targeted supervision settings, which we hypothesize to be a natural capability of PFNs. In this work, we also adopt the PFN paradigm for tabular anomaly detection but with a focus on in-context adaptation across supervision regimes.

\section{In-Context Learning for Tabular AD}

\subsection{Problem Formulation}

We begin by reformulating tabular anomaly detection as a task-level inference problem. We are given a training/context dataset $\mathcal{C} = (\mathcal{X}, \mathcal{Y})$, where
$\mathcal{X} = \{x_i\}_{i=1}^N$ denotes tabular samples that may contain an unknown but typically small proportion of anomalies and $\mathcal{Y} = \{y_i\}_{i=1}^N$ denotes labels with $y_i \in \{0,1,\varnothing\}$, where 0 indicates normal samples, 1 indicates anomalous samples, and $\varnothing$ denotes the absence of a label. The goal is to infer a context-conditioned anomaly scoring function $s(\cdot \mid \mathcal{C}) : \mathbb{R}^d \rightarrow \mathbb{R}$, which, when applied to unseen (query) samples, assigns higher scores to anomalous samples than to normal samples.

This formulation unifies common supervision regimes in tabular anomaly detection as special cases of the same task-level inference problem. In the \emph{one-class} setting, all samples are assumed to be normal and labeled as such, i.e., $y_i = 0$ for all $i$. In the \emph{unsupervised} setting, labels are completely absent ($y_i = \varnothing$ for all $i$), and the context may contain an unknown proportion of anomalous samples. In the \emph{semi-supervised} setting, a small subset of anomalies are labeled as anomalous ($y_i = 1$), while the remaining samples are unlabeled. Importantly, in all cases, the absence of a label does not imply normality, and neither the supervision regime nor the contamination level is assumed to be known to the model \emph{a priori}, i.e. before observing the training set.

\subsection{Context-Conditioned Anomaly Scoring}

We interpret anomaly detection as estimating the probability that a query sample is anomalous given a context dataset.
Concretely, given a dataset $\mathcal{C}$, we define the anomaly score of a query sample $x$ as the conditional probability
\[
s(x \mid \mathcal{C}) = p(y = 1 \mid x, \mathcal{C}),
\]
This formulation conditions the anomaly score on both the distributional patterns present in the dataset and any supervision available in the context.

From a probabilistic perspective, this conditional probability can also reflect the uncertainty over the underlying data-generating process.
Let $\phi$ denote a latent data-generating mechanism drawn from a prior
distribution $\Pi$.
Conditioning on the context induces a posterior over plausible
mechanisms, and the anomaly score can be interpreted as a posterior predictive distribution (PPD):
\begin{equation}
p(y = 1 \mid x, \mathcal{C})
= \int_{\Pi} p(y = 1 \mid x, \phi)\, p(\phi \mid \mathcal{C})\, d\phi.
\label{eq:ano_ppd}
\end{equation}

Although this probability is generally intractable to compute, it motivates an approach that approximates the mapping from $(x, \mathcal{C})$ to $p(y=1 \mid x, \mathcal{C})$ directly. PFNs learn this mapping by training over a large number of synthetic tasks sampled from the prior distribution. During inference, a PFN estimates the PPD by in-context learning, and thus approximates Bayesian inference over data-generating mechanisms of normal and anomalous samples. 

\begin{figure}[t]
  \centering
  \includegraphics[width=\linewidth]{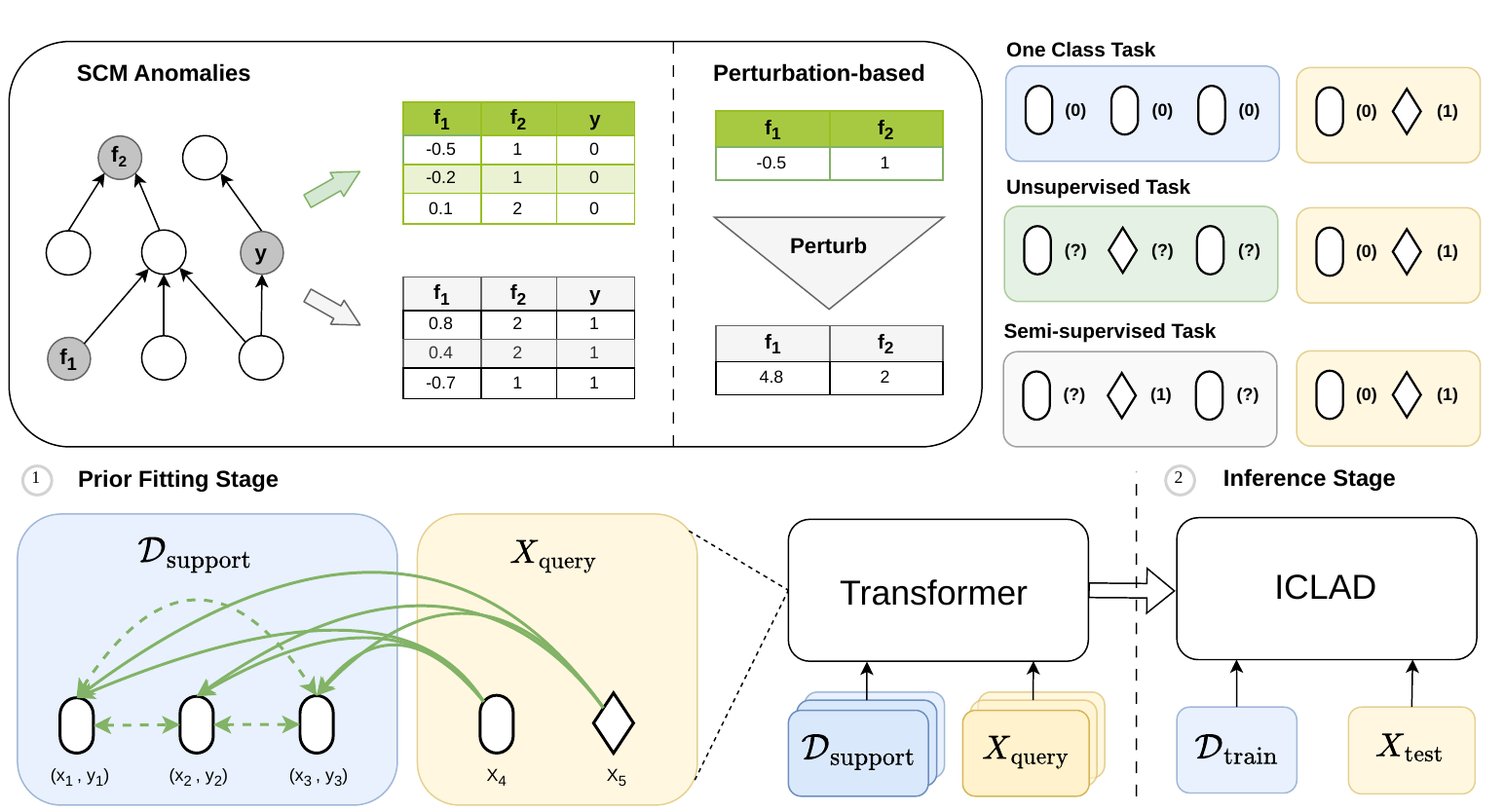}
  \caption {
    Overview of the ICLAD framework and synthetic task construction. The top region depicts anomaly generation and the construction of supervision tasks. The bottom region shows the two-stage procedure of ICLAD. Left: prior-fitting on synthetic tasks. Right: inference on real datasets.
    }
  \label{ICLAD_overview}
\end{figure}

\section{ICLAD Framework}

We propose ICLAD, an in-context learning framework for tabular anomaly detection that unifies one-class, unsupervised, and semi-supervised settings using a single model trained on a distribution of anomaly tasks. ICLAD follows a two-stage procedure consisting of a prior-fitting stage and an inference stage.

\paragraph{\textbf{Prior-Fitting Stage}} In this stage, we train an in-context learning model $q_\theta$ parameterized by weights $\theta$ on synthetic anomaly detection tasks sampled from a distribution $p(\mathcal{T})$. Each task $\mathcal{T}$ consists of a support set \( \mathcal{D}_{\text{support}} \) and a query set \(\mathcal{D}_{\text{query}} = \{(x_j, y_j)\}_{j=1}^M\). The support set defines the context, which conditions the model’s predictions for query samples. The model parameters $\theta$ are optimized by minimizing the binary classification error on the query samples using the following objective:
\begin{displaymath}
\mathcal{L}_{BCE} = \mathbb{E}_{\mathcal{T} \sim p(\mathcal{T})} \left[ \text{BCE}\left(y_j, q_\theta(x_j \mid \mathcal{D}_{\text{support}}) \right) \right],
\label{eq:adpfn_bce}
\end{displaymath}
where BCE denotes the binary cross-entropy loss, and the expectation is taken over a large number of tasks sampled from $p(\mathcal{T})$. After training, the model $q_{\theta}$ effectively approximates the anomaly scoring as defined in Equation \ref{eq:ano_ppd}.

\paragraph{\textbf{Inference Stage}} At inference time, the learned parameters $\hat{\theta}$ are fixed. Given a real dataset $\mathcal{D}_{\text{train}}$, we treat it as the support set and compute the context-conditioned anomaly scores for new samples using the model $q_{\hat{\theta}}(\cdot \mid \mathcal{D}_{\text{train}})$. ICLAD can thus adapt its scoring behavior based on context data, requiring only forward computations at test time. Notably, the predictions for query samples are conditionally independent given the support set and are not influenced by other query instances.

\subsection{Synthetic Anomaly Detection Tasks}

To enable generalization across tabular anomaly detection tasks with diverse data characteristics and supervision settings, we construct a synthetic task distribution that reflects this variability. Task generation consists of three stages: (i) sampling tabular datasets from structural causal models (SCMs) (ii) generating anomalies through structural and perturbation-based mechanisms, and (iii) constructing tasks under one-class, unsupervised and semi-supervised regimes.

\paragraph{\textbf{Simulating Tabular Datasets}}
We generate synthetic tabular datasets using the structural causal model (SCM) prior introduced in TabPFN~\cite{hollmannTabPFNTransformerThat2022a}. An SCM defines a joint distribution over variables through a directed acyclic graph (DAG) $G_{\text{scm}}$, where each variable $z_i$ is generated by a causal mechanism
$
z_i = f_i(z_{\mathrm{pa}(i)}, \epsilon_i),
$
with $\mathrm{pa}(i)$ denoting the parents of node $i$, $f_i$ a deterministic function, and $\epsilon_i$ a Gaussian noise term. 

Synthetic datasets are obtained by sampling from the SCM and selecting a subset of variables as observed features. One variable is discretized to form a binary label $y \in \{0,1\}$, which induces class-conditional distributions $p_{\text{scm}}(x \mid y)$ that correspond to normal ($y=0$) and anomalous ($y=1$) samples. To emulate real tabular data, a random subset of features is discretized to produce mixed continuous and categorical variables. This process creates datasets with varying dimensionality, heterogeneous feature types, and complex feature relationships.

\begin{figure}[t]
    \centering
    \begin{subfigure}[b]{0.495\textwidth}
        \centering
        \includegraphics[width=0.49\textwidth]{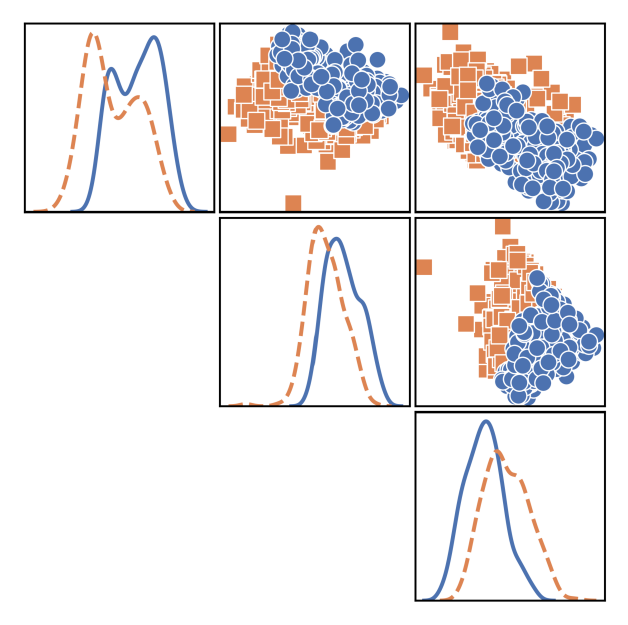}
        \includegraphics[width=0.48\textwidth]{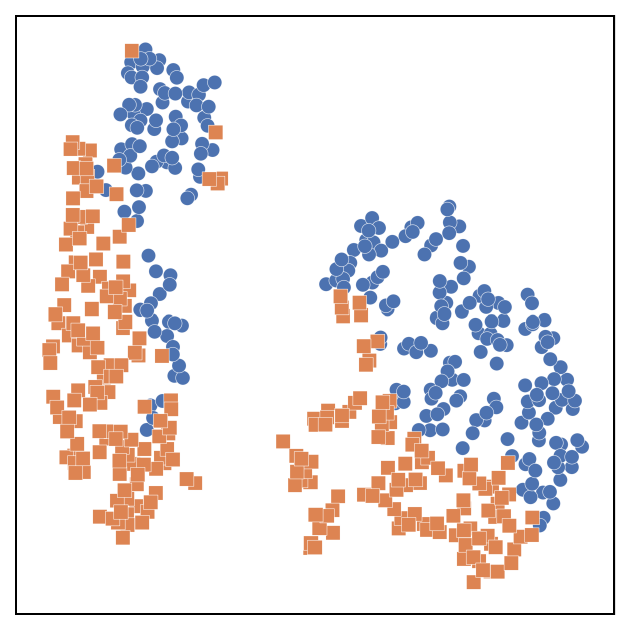}
        \caption{SCM Anomalies}
        \label{fig:scm_anom_query_set}
    \end{subfigure}
    \hfill
    \begin{subfigure}[b]{0.495\textwidth}
        \centering
        \includegraphics[width=0.49\textwidth]{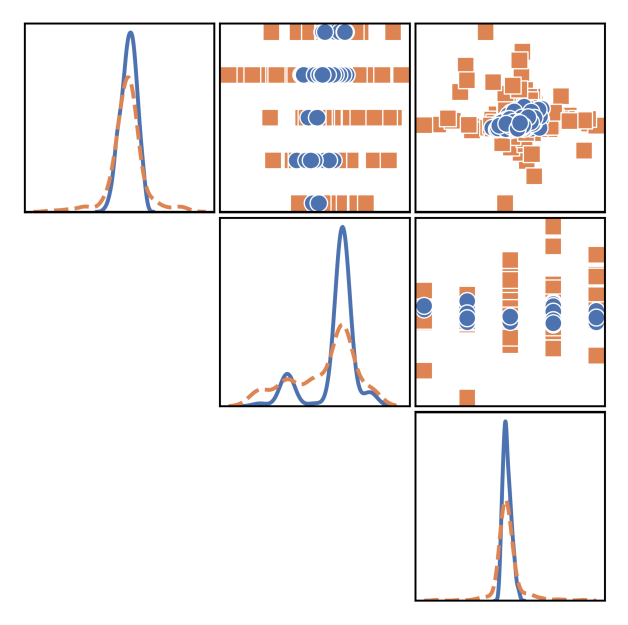}
        \includegraphics[width=0.48\textwidth]{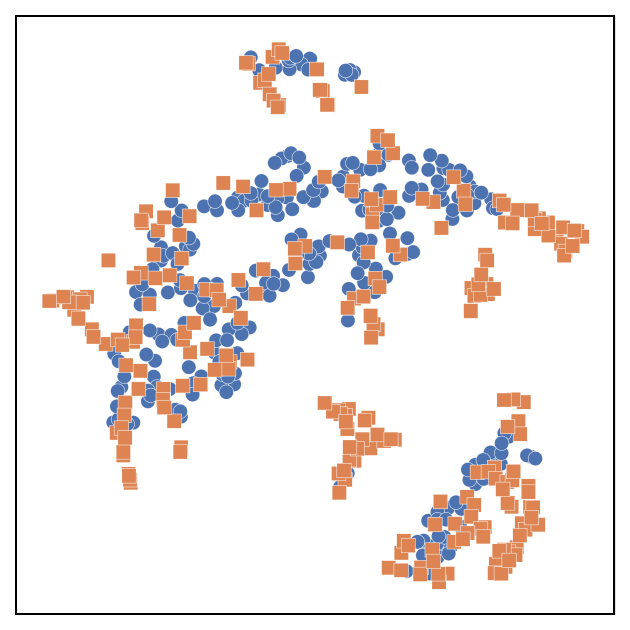}
        \caption{Perturbation-based Anomalies}
        \label{fig:pert_anom_query_set}
    \end{subfigure}
    \caption{Feature interaction and t-SNE~\cite{maatenVisualizingDataUsing2008} plots of normal and anomalous samples. The orange rectangles are anomalies; the blue circles are normal samples.}
    \label{fig:anom_visualizations}
\end{figure}

\paragraph{\textbf{Simulating Anomalies}}
To model the diversity of anomalies encountered in practice, we employ two complementary anomaly generation mechanisms. The first mechanism produces \emph{structural anomalies} by sampling from the anomalous class generated by the SCM, $p_{scm}(x \mid y=1)$, resulting in samples that exhibit global distributional shifts relative to the normal data, which are sampled from the distribution $p_{scm}(x \mid y=0)$. These anomalies reflect shifts in the underlying causal mechanisms and capture structured, semantic deviations relative to normal samples as shown in Figure~\ref{fig:scm_anom_query_set}. Crucially, these samples share an underlying relational structure which, when labeled, can create meaningful supervision to guide anomaly scoring in semi-supervised setting.  

The second mechanism generates \emph{perturbation-based anomalies} by corrupting selected features of normal samples. For continuous features, we sample a feature-wise sparsity mask from a Bernoulli distribution for each sample and apply additive Gaussian noise,
\[
\tilde{x}_i = x_i + m_i \sigma_i \epsilon_i, \quad \epsilon_i \sim \mathcal{N}(0,1),
\]
where $m_i \in \{0,1\}$ indicates whether feature $i$ is perturbed and $\sigma_i$ is drawn independently per sample and feature from a log-uniform distribution. For categorical features, corruption is implemented by randomly replacing the original value with another valid category.

\paragraph{\textbf{Constructing Supervision Tasks}}

A central component of ICLAD is the construction of synthetic anomaly detection tasks which span the commonly encountered supervision regimes. Each task consists of a support or the context set $\mathcal{D}_{\text{support}}=(X_{\text{support}},Y_{\text{support}})$ and a query set $\mathcal{D}_{\text{query}}=(X_{\text{query}},Y_{\text{query}})$. Query labels satisfy $Y_{\text{query}} \in \{0,1\}$, while support labels satisfy $Y_{\text{support}} \in \{-1,0,1\}$, where $-1$ denotes unlabeled samples. The query set is constructed as a balanced mixture of normal and anomalous samples with equal proportions of structural and perturbation-based anomalies.

Contrary to the query set, the composition and labeling of the support set vary across tasks to simulate different supervision regimes:

\begin{enumerate}
    \item In the \emph{one-class} setting, the support set contains only normal samples drawn from $p_{\text{scm}}(x \mid y=0)$, and all support labels are $0$. 
    \item In the \emph{unsupervised} regime, the support set is unlabeled and contaminated with anomalies. The anomaly ratio is sampled uniformly from $[0,0.4]$ and all labels in $Y_\text{support}$ are $-1$, indicating unknown.
    \item In the \emph{semi-supervised} regime, the support set is constructed as in the unsupervised setting, but a uniformly sampled fraction of anomalous samples is labeled as $1$, while all remaining samples remain -1.
\end{enumerate}

Across all regimes, all anomalous samples in the support set are drawn from the structural anomaly class induced by the SCM. Perturbation-based anomalies are applied in the query set which complements the structural anomalies and guide the model to detect anomalies beyond the support of labeled anomalies, i.e. open-set anomalies. During the prior-data fitting, we uniformly sample tasks from the \emph{one-class}, \emph{unsupervised}, and \emph{semi-supervised} regimes, exposing the model to diverse supervision scenarios and enabling it to infer the supervision setting from its context. The resulting joint distribution over SCMs, supervision regimes, and anomaly generation mechanisms defines the prior we designed for tabular anomaly detection.

\subsection{Model Architecture}

ICLAD adopts the transformer architecture of TabPFN~\cite{hollmannTabPFNTransformerThat2022a}, where support and query samples are processed jointly using an asymmetric attention pattern that prevents information leakage between queries and implements $q_\theta(\cdot \mid \mathcal{D}_{support})$. Input samples are zero-padded to a fixed dimension of 512 and projected to the model embedding space, and labels are embedded using a lookup table $E_y$ with unlabeled samples mapped to the zero vector.

\paragraph{FiLM Label Conditioning}
To incorporate supervision signals in the support set, we apply feature-wise linear modulation (FiLM)~\cite{perezFiLMVisualReasoning2018} to support sample embeddings. Given an embedding $x$ and its label embedding $c = E_y(y)$, FiLM computes
\[
\tilde{x} = (1+\gamma(c)) \odot x + \beta(c),
\]
where $\gamma(\cdot)$ and $\beta(\cdot)$ are learned linear mappings. Unlabeled samples correspond to $c=0$, for which FiLM reduces to the identity transformation.

\subsection{Key Implementation Details}

In this section, we provide some key implementation details and defer the complete details to section A of the supplementary materials.

\paragraph{Training} We instantiate ICLAD as a 12-layer Transformer~\cite{vaswaniAttentionAllYou2017a} trained over 52 million tasks. Training takes about 53 hours over 4 Nvidia H100 GPUs.

\paragraph{Context Size and Feature Size} During the prior fitting, the number of support samples is sampled uniformly from 5 to 12{,}000, with feature dimensions ranging from 2 to 512, which enables ICLAD to be directly applicable on moderate sized tabular datasets. Adapting to larger datasets or those with higher dimensionality is enabled through feature and context subsampling.

\paragraph{Ensembling}  Following standard PFN implementations, predictions are averaged across multiple ensemble tasks, each operating on different subsets of support samples and feature ordering or feature subsets. This generally improves performance and makes predictions more invariant to feature permutations. 

\paragraph{KV Caching} Since the predictions are conditionally independent given the support set, we decouple the processing of context samples from query evaluation using key-value (KV) caching. This allows support representations to be computed once and reused across query batches and enables an efficient two-stage fit-and-predict pipeline.

\section{Experiments}

\subsection{Setup} We evaluate ICLAD on the ADBench benchmark suite, which contains 57 real-world anomaly detection tabular datasets. These datasets cover a wide range of data characteristics and anomaly ratios, to illustrate, dimensionality ranges from 3 to up to 1{,}555, and anomaly ratios vary from approximately 3\% to 39\%. 

Experiments are conducted under the one-class, unsupervised, and semi-supervised settings. In the one-class setting, the training dataset is constructed by sampling 50\% of the normal samples from the full dataset without replacement. The remaining normal samples together with all anomalous samples form the test set.
In the unsupervised setting, we follow the protocol proposed by DTE~\cite{livernocheDiffusionModelingAnomaly2023b}, where the training set is obtained by bootstrapping for robustness evaluation, and the full dataset is used for testing. In the semi-supervised setting, datasets are partitioned into 70\% training and 30\% test data using stratified sampling to preserve the original anomaly ratio. Within the same training data, a fraction $r_a$ of anomalous samples is randomly selected and assigned anomaly labels, while all remaining samples remain unlabeled.

We evaluate all methods using three standard anomaly detection metrics: AUC-ROC, AUC-PR, and F1. Due to space constraints, we report mainly AUC-ROC, while AUC-PR and F1 results are provided in section E of the supplementary. The statistical significance of model performance are assessed by critical difference (CD) diagrams based on the Friedman test with Wilcoxon-Holm post-hoc analysis at a significance level of 0.05~\cite{demsarStatisticalComparisonsClassifiers2006}. All experiments are repeated with five random seeds, and results are reported as the average performance over the seeds and datasets.

To maintain reasonable computation time, larger datasets are subsampled to a maximum of 100{,}000 samples. This is substantially larger than the 10{,}000 sample cap used in previous works~\cite{hanADBenchAnomalyDetection2022a,livernocheDiffusionModelingAnomaly2023b}, allowing us to evaluate the scalability of ICLAD on larger datasets.

\subsection{Baseline Methods}

\paragraph{Classical Baselines} 
We compare ICLAD against several widely used classical baselines, including CBLOF, ECOD, iForest, kNN, LOF, OCSVM, PCA, MCD, HBOS, 
and the non-parametric version of DTE, DTE-NP.

\paragraph{Deep Learning Baselines} For deep learning models, we include Autoencoder (AE), Deep-SVDD, SLAD, ICL, GOAD, MCM, DRL, NeuTraLAD (NTL), LUNAR, Fomo-0D, and the parametric DTE variants, DTE-C and DTE-IG. For the unsupervised learning scenario, we include LOE with the NTL backbone. 

\paragraph{Semi-Supervised Baselines} 
We select well known semi-supervised approaches such as DevNet, FeaWAD, DeepSAD, PreNet and GANomaly to measure how well ICLAD utilize available labels.

For hyperparameters, we use the default PyOD~\cite{chenPyOD2Python2025} configurations where applicable. Other methods follow their recommended settings.

\section{Results}

\subsubsection{One-class and Unsupervised Regimes}

\begin{figure}[t]
    \centering

    \begin{subfigure}{0.85\textwidth}
        \centering
        \includegraphics[width=\textwidth]{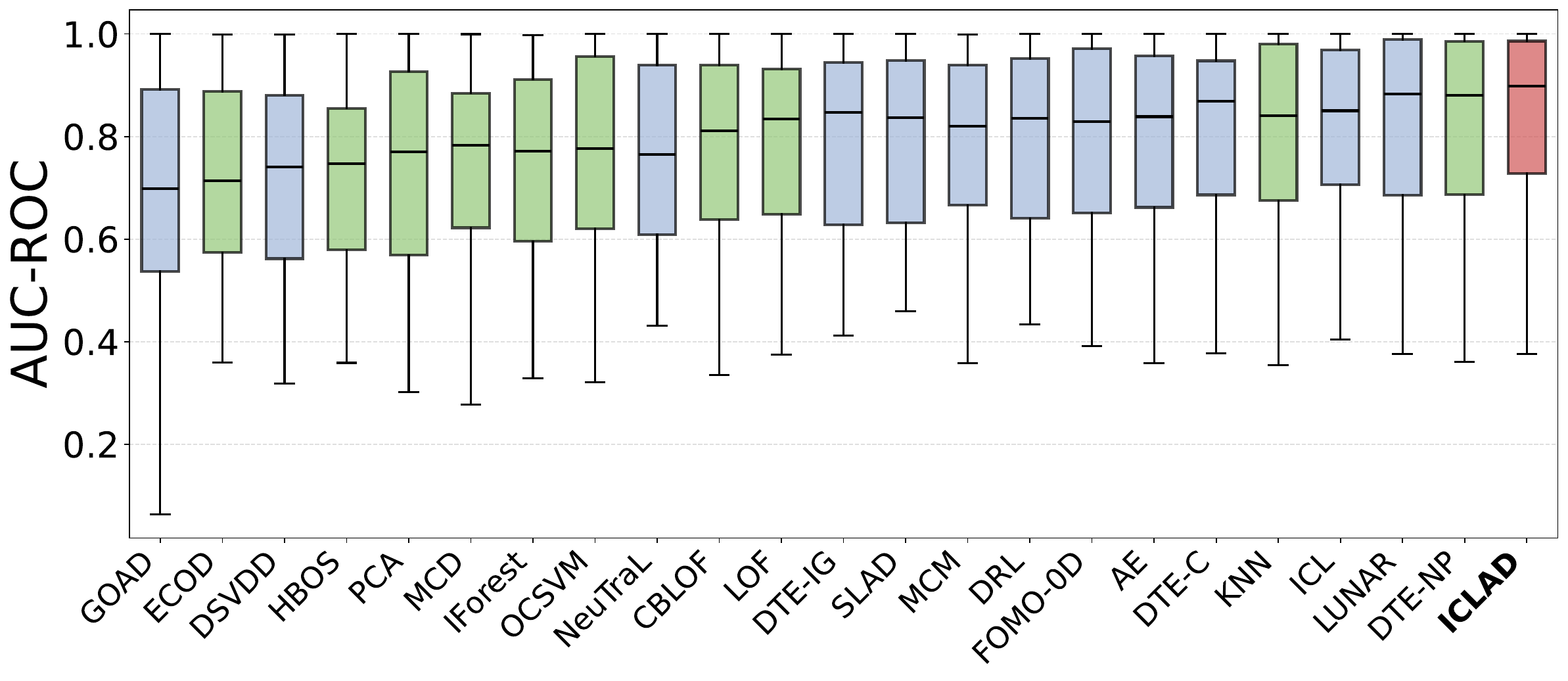}
        \caption{One-class setting}
    \end{subfigure}

    \begin{subfigure}{0.85\textwidth}
        \centering
        \includegraphics[width=\textwidth]{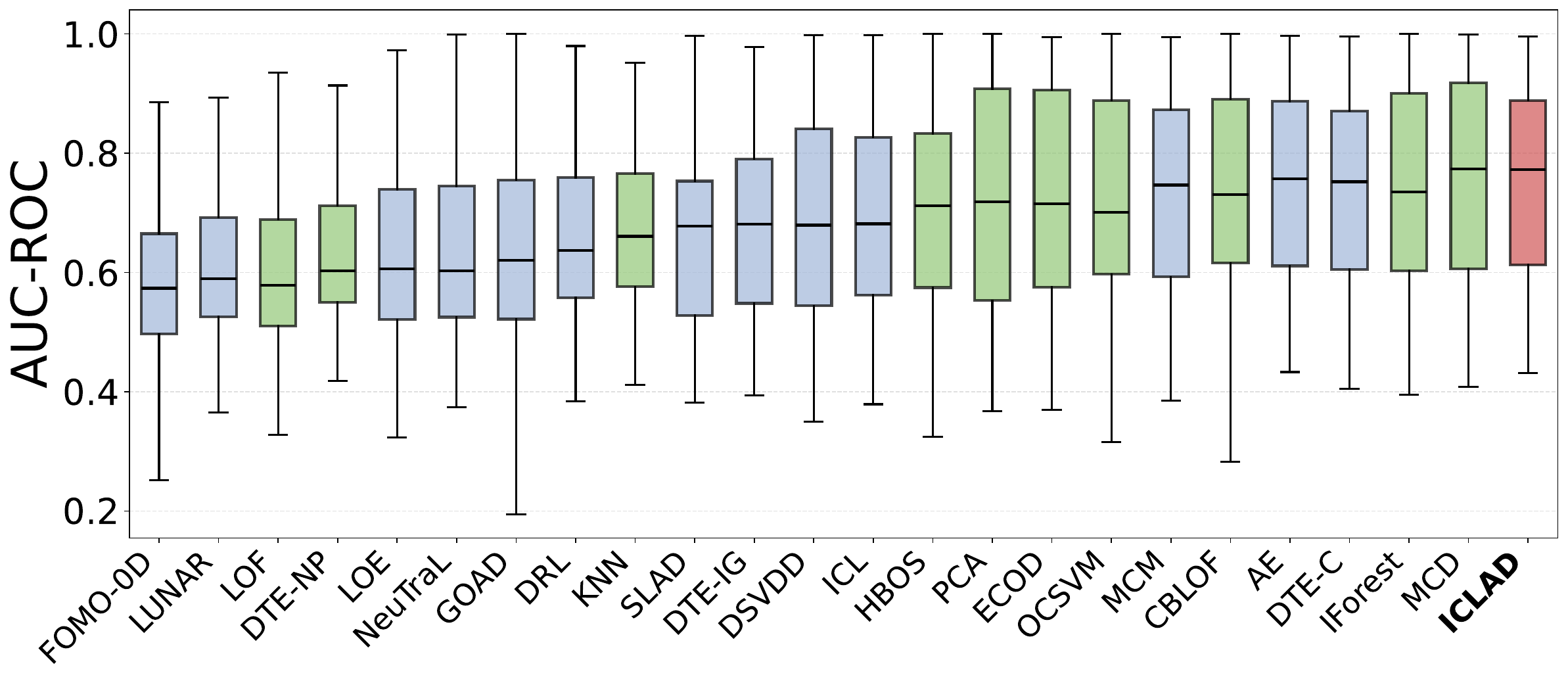}
        \caption{Unsupervised setting}
    \end{subfigure}

    \caption{Boxplots of AUC-ROC across 57 datasets. Boxes show the interquartile range (IQR) with medians indicated by the center line and whiskers extending to 1.5 times IQR. Models are ordered by average AUC-ROC and color-coded by family: classical (green), deep learning (blue), and ICLAD (red).}
    \label{fig:iclad_boxplots_main}
\end{figure}

\begin{figure}[t]
    \centering

    \begin{subfigure}[a]{0.75\textwidth}
        \centering
        \includegraphics[width=\linewidth]{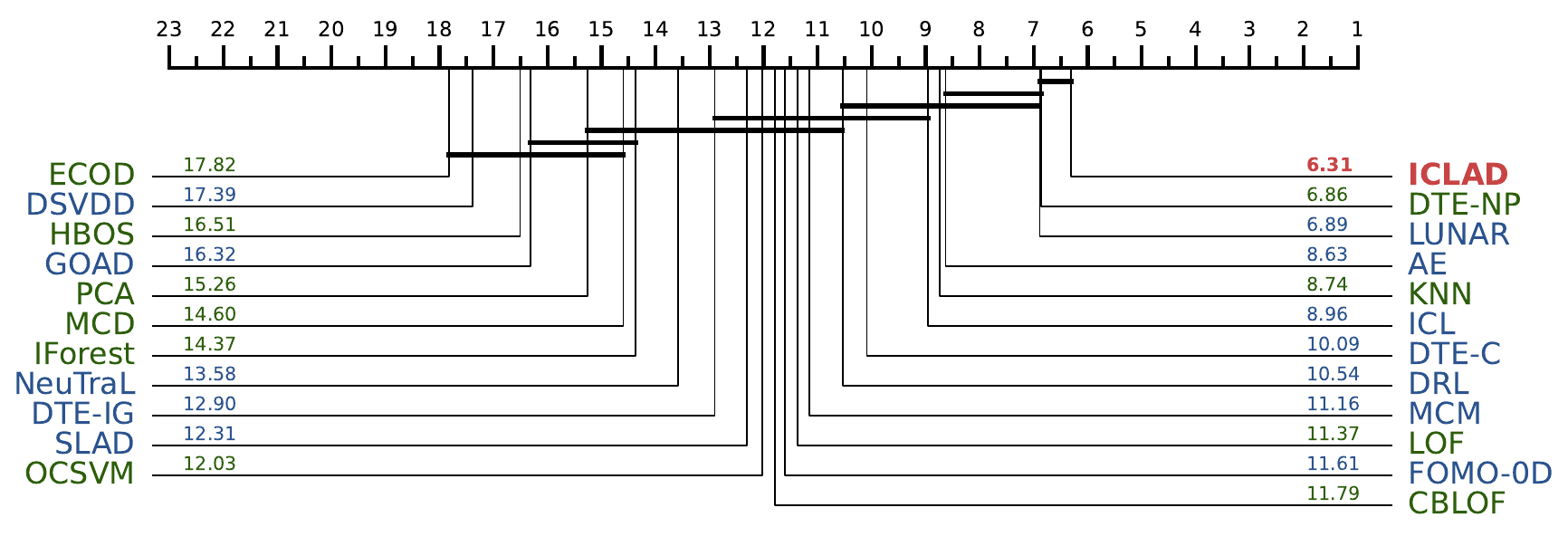}
        \caption{One-class setting}
        \label{fig:iclad_oc_cdd_aucroc}
    \end{subfigure}

    \begin{subfigure}[b]{0.75\textwidth}
        \centering
        \includegraphics[width=\linewidth]{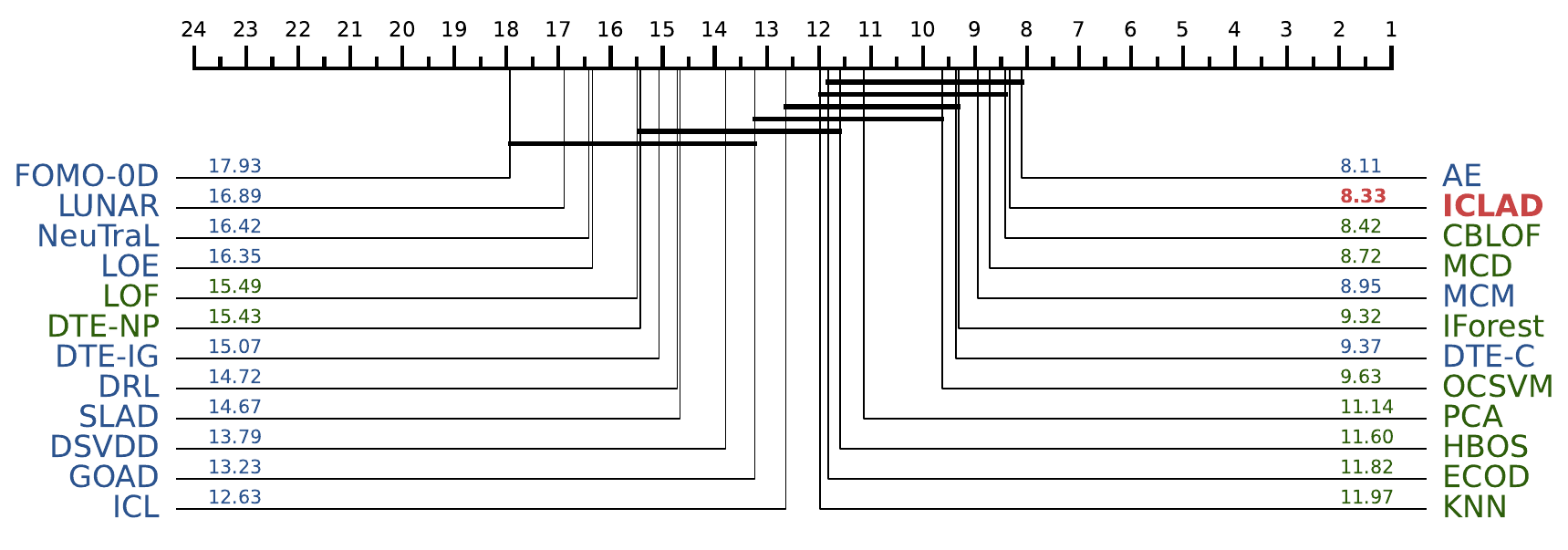}
        \caption{Unsupervised setting}
        \label{fig:iclad_unsup_cdd_aucroc}
    \end{subfigure}

    \caption[ICLAD One-Class Setting Critical Difference Diagrams]{
        Critical difference diagrams of average AUC-ROC ranks. Models are color coded by: classical (green), deep learning (blue) and ICLAD (red)
    }
    \label{fig:iclad_cdds_main}
\end{figure}

Figure~\ref{fig:iclad_boxplots_main} reports the performances in the one-class and unsupervised setting.
ICLAD achieves the highest average AUC-ROC among all baselines which shows strong generalization to real world datasets. In the one-class setting, ICLAD achieves decent improvement gains over prior arts with wider performance gaps in AUC-PR and F1 scores. This is further supported by the CD-diagram in Figure~\ref{fig:iclad_cdds_main}, showing that ICLAD obtains the lowest average rank. Notably, DTE-NP, a variant of kNN, performs strongly in this regime and appears in the same top-performing clique as ICLAD under the Wilcoxon–Holm post-hoc test. This is consistent with prior observations that local-density methods are highly competitive under the one-class setting~\cite{livernocheDiffusionModelingAnomaly2023b,shenkarAnomalyDetectionTabular2021a}.

In the unsupervised regime, ICLAD's performance is on par with contamination robust methods such as MCD and strong classical baselines including CBLOF and iForest, as illustrated in the boxplot and CD diagram. While we do not observe significant differences in performance compared to these baselines, it is important to note that strong performing methods in the one-class setting are generally less robust under this setting. For example, DTE-NP and LUNAR exhibit large performance drops due to their sensitivity to contamination in the training data. In contrast, ICLAD demonstrates greater robustness to contamination and is the only method that consistently ranks among the top performers across both the one-class and unsupervised settings.

\begin{figure}[h]
    \centering
    \includegraphics[width=\linewidth]{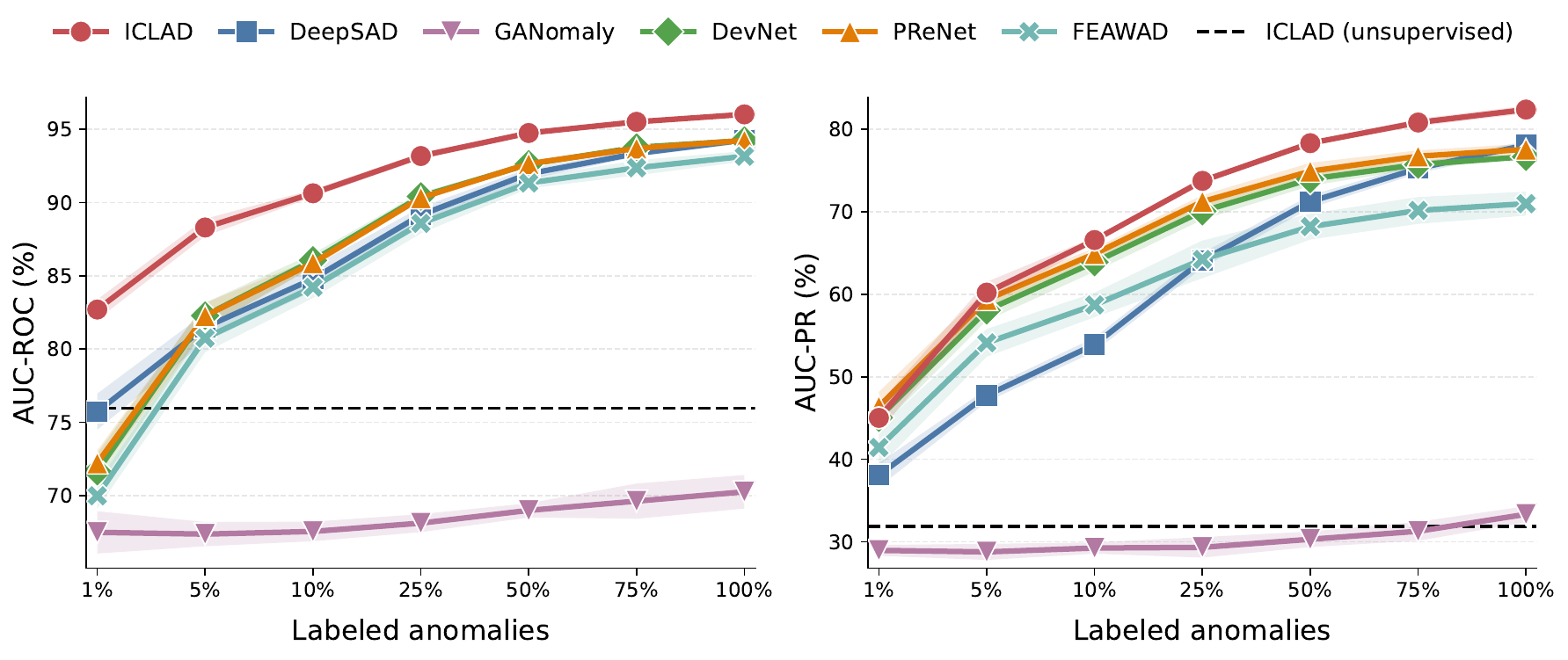}

    \caption[Effect of Labeled Anomaly Availability on ICLAD in the Semi-supervised Setting]{
        Semi-supervised performance curves against ratio of labeled anomalies. Left: AUC-ROC Right: AUC-PR
    }
    \label{fig:iclad_semi_main}
\end{figure}

\subsubsection{Semi-supervised Regime} In the semi-supervised setting, where partial supervision is available through labeled anomalies, substantial performance improvements are observed across all methods, as shown in Figure~\ref{fig:iclad_semi_main}. With 5\% labeled anomalies, nearly all semi-supervised baselines outperform ICLAD in the unsupervised setting, showcasing the benefit of label information. Notably, ICLAD maintains a performance margin over all baselines in both the AUC-ROC and AUC-PR curves. Furthermore, with only 10\% labeled anomalies, ICLAD achieves a 10\% increase in AUC-ROC compared to its unsupervised counterpart, demonstrating that the model benefits significantly from partial supervision.

\begin{figure}[!h]
\centering

\begin{minipage}[!h]{0.52\textwidth}
\centering
\includegraphics[width=\linewidth]{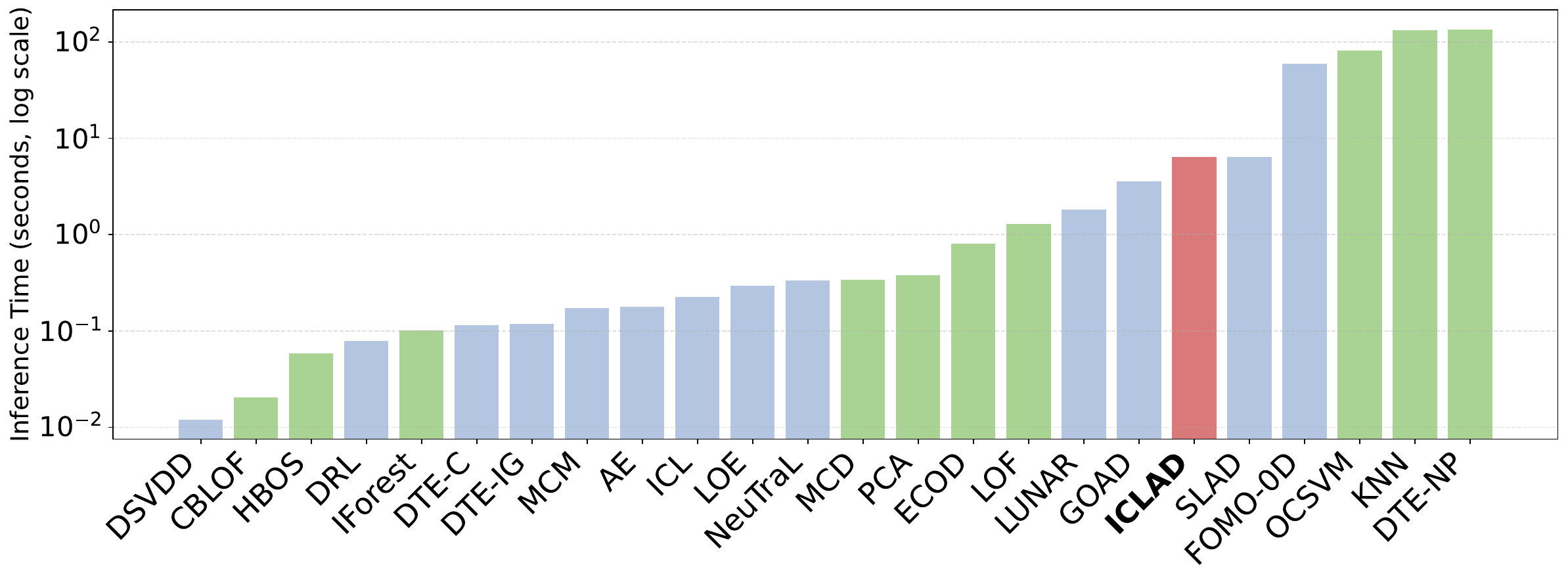}

\includegraphics[width=\linewidth]{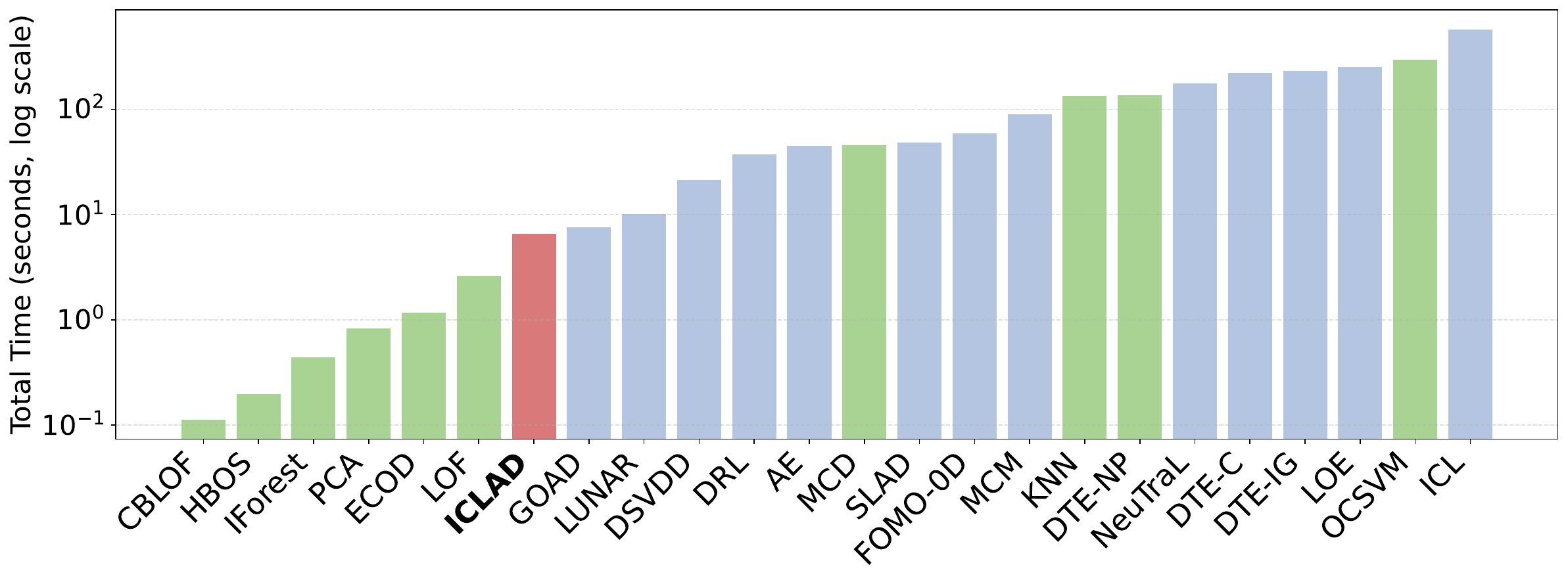}
\end{minipage}
\hfill
\begin{minipage}[t]{0.47\textwidth}
\centering
\footnotesize
\setlength{\tabcolsep}{3pt}
\begin{tabular}{lcc}
\toprule
Variants & One-class & Unsup. \\
\midrule
One-class only & \textbf{84.00} & 68.26\\
Unsup. only & 82.61 & 74.80\\
SCM only & 83.37 & 74.14 \\
\midrule
ICLAD & 83.97 & \textbf{75.17}\\
\bottomrule
\end{tabular}
\end{minipage}

\caption{Efficiency and ablation analysis. Left Top: average inference runtime (seconds). Left Bottom: average total runtime (training + inference)(seconds). Runtime axes use a log scale. Right: Ablation study of ICLAD variants. Performance are measured in average AUC-ROC across the respective supervision setting.} 
\label{fig:efficiency_ablation}
\end{figure}

\subsection{Ablation and efficiency analysis}

For the ablation study, we evaluate variants of ICLAD trained with only unsupervised tasks, only one-class tasks, and a version using only SCM-based anomalies. As expected, the one-class-only model performs best in the one-class setting, yet the model degrades substantially on the unsupervised benchmark as shown on the right of Figure~\ref{fig:efficiency_ablation}. Our model has the most balanced performance out of the variants over one-class and unsupervised benchmarks. This justifies the inclusion of unsupervised tasks as well as perturbation-based anomalies in the query set.

In terms of efficiency, ICLAD exhibits moderate inference time due to the quadratic complexity of self-attention. Nevertheless, when considering the total runtime (training and inference), it remains competitive and is only outperformed by traditional baselines as shown on the left of Figure~\ref{fig:efficiency_ablation}. All runtime measurements were conducted on a single NVIDIA H100 GPU.

\section{Conclusion}

We introduced ICLAD, an in-context learning framework for anomaly detection. The emperical results show ICLAD acquires inductive bias aligned to real-world anomaly detection problems and achieves consistently competitive performance in all three supervision settings. Our findings suggest that in-context learning provides a promising direction for more general anomaly detection systems. Future works can focus on developing better synthetic task generation strategies that improves generalization especially under the case of anomaly contamination.



%
%

%
\bibliographystyle{splncs04}
\bibliography{references}
%
\newpage
\appendix

\section{Additional Implementation Details}

\subsection{Model Architecture}
We adapt the Transformer encoder architecture \cite{vaswaniAttentionAllYou2017a} from TabPFN \cite{hollmannAccuratePredictionsSmall2025b} in our implementation of ICLAD. The transformer encoder consists of 12 layers, each with 4 attention heads, a hidden dimension of 512, and a feed-forward network with an intermediate dimension of 1024. The feed-forward networks employs the GELU activation function \cite{hendrycksGaussianErrorLinear2023a}. We adopt a post-normalization architecture, following the original Transformer implementation, but omit positional encodings to respect the permutation invariance of independent tabular samples. The model contains approximately 26.5 million trainable parameters.

\subsection{Input Preprocessing}

All features are standardized using z-score normalization and clipped to the range [-100, 100] to improve numerical stability during optimization. The ICLAD model assumes that all inputs are preprocessed in this manner. For each task, consisting of a support set and a query set, normalization is performed using the mean and standard deviation computed from the support set only. Specifically, for each feature $j$, we compute the mean \( \mu_j \) and standard deviation \( \sigma_j \) over the support set, and normalize each sample's $j$th dimension \( x \) as
\[
\tilde{x}_j = \frac{x_j - \mu_j}{\sigma_j}.
\]
These statistics are then applied to both the support and query samples to prevent information leakage from the query set. Note that the z-score normalization is applied uniformly to all features, without distinguishing between continuous and discrete variables. This feature-type agnostic design aligns with previous anomaly detection approaches, which typically operate directly on vectorized features without specialized embeddings for categorical or ordinal variables. Following established protocols, we embed inputs as a vector, and all features are treated uniformly by the model. Features with zero variance are left unchanged, following the behavior of standard Scikit-learn implementation~\cite{pedregosaScikitlearnMachineLearning2011}.

The maximum number of features admissible for the ICLAD transformer, \( d_{\max} \), is set to 512, corresponding to the model hidden dimension. Input samples with fewer than \( d_{\max} \) features are zero-padded to \( d_{\max} \), following TabPFN.

To ensure consistent input magnitudes across datasets with varying numbers of features \( d_{\text{in}} \), we rescale each zero-padded input vector \( \tilde{x} \in \mathbb{R}^{d_{\text{max}}} \) as
\[
x = \sqrt{\frac{d_{\max}}{d_{\text{in}}}} \, \tilde{x}.
\]
Under the assumption that features are standardized, this scaling normalizes the expected squared \( \ell_2 \)-norm of the input vectors, making it approximately invariant to \( d_{\text{in}} \). This normalization step helps stabilize training and improves generalization across tabular datasets with varying dimensionality.

\subsection{Training and Optimization Details}

The ICLAD model is trained for 100 epochs during the prior fitting stage using the Adam optimizer with no weight decay. Training is distributed across 4\(\times\) NVIDIA H100 GPUs using mixed precision.

Each epoch consists of 2048 training steps. At each step, every GPU processes 16 tasks, resulting in a per-step batch size of 64 across all GPUs. We employ gradient accumulation over 8 steps, resulting in an effective batch size of 512 with 256 optimization steps per epoch. A learning rate schedule with linear warm-up (3 epochs) followed by cosine annealing is used, with a maximum learning rate of \(1 \times 10^{-4}\).

In total, ICLAD is trained on \(13{,}107{,}200\) synthetic anomaly detection tasks. Training takes approximately 55 wall-clock hours per GPU, corresponding to about 220 GPU hours in total.

\subsection{Key-Value Caching}

We employ key-value (KV) caching to accelerate both training and inference by exploiting the asymmetric structure of in-context learning. In particular, query samples attend only to support samples, while support samples do not attend to queries. This allows the in-context learning inference to be decomposed into two stages.

\paragraph{Fitting stage}
Given a support set $X_s$, we compute and cache the key and value representations at each layer:
\[
(K_s^{(\ell)}, V_s^{(\ell)}) = \mathrm{KVProj}^{(\ell)}(X_s), \quad \ell = 1, \dots, L.
\]

\paragraph{Predict stage}
Given a query set $X_q$, we compute query representations and attend only to the cached support keys and values:
\[
\mathrm{Attention}^{(\ell)} = \mathrm{softmax}\left( \frac{Q_q^{(\ell)} (K_s^{(\ell)})^\top}{\sqrt{d}} \right) V_s^{(\ell)}.
\]

Compare to using full attention with specialized masks as in TabPFN~\cite{hollmannTabPFNTransformerThat2022a}, this avoids recomputing support representations for each query and reduces redundant attention operations, leading to improved efficiency.

\section{Additional Details of Synthetic Prior Task Generation}

In this section, we describe the construction of synthetic prior tasks used to train ICLAD. Following the prior-data fitted networks (PFN) framework, tasks are created using data-generating processes that define both the input distribution and the associated prediction objective. These processes are based on structural causal models (SCMs) and perturbation-based corruption mechanisms, each parameterized by task-specific hyperparameters.

\begin{figure}[t]
    \centering
    \begin{subfigure}[b]{0.38\textwidth}
        \centering
        \includegraphics[width=\linewidth]{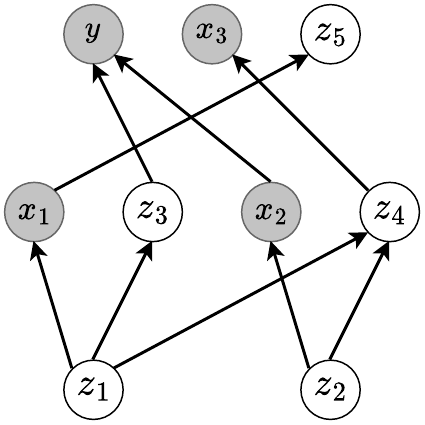}
        \caption{SCM Example 1}
        \label{fig:scm_1}
    \end{subfigure}
    \hspace{2cm}
    \begin{subfigure}[b]{0.27\textwidth}
        \centering
        \includegraphics[width=\linewidth]{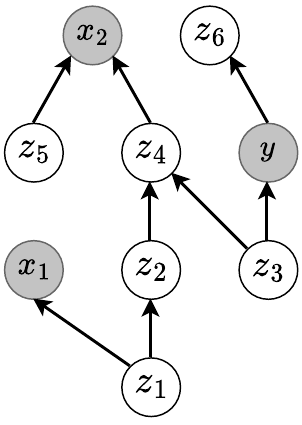}
        \caption{SCM Example 2}
        \label{fig:scm2}
    \end{subfigure}
    
    \caption{Two visual examples of SCM that can be sampled from the SCM prior. Gray circles represent observed variables and white circles are latent variables. \(x_i\) are selected features and \(y\) is the (continuous) target variable. }
    \label{fig:SCM_examples}
\end{figure}

\subsection{SCM Data Generation}

While TabPFN considers a mixture of structural causal models (SCMs) and Bayesian neural networks (BNNs) during prior-data fitting, we adopt only the SCM prior. This choice is motivated by prior ablations showing that the SCM prior alone is sufficient to achieve strong performance across a wide range of tabular tasks, suggesting better alignment with real-world tabular data.

We briefly describe the SCM generation process (see Appendix C.1 of \cite{hollmannTabPFNTransformerThat2022a} for full details). A directed acyclic graph (DAG) is first sampled to define dependencies among variables:

\begin{enumerate}
    \item Sample the architecture of a multi-layer perceptron (MLP), including the number of layers \( \ell \sim p(\ell) \) and hidden nodes per layer \( h \sim p(h) \).
    \item Initialize network weights using normal distributions and randomly drop connections to induce sparsity, resulting in a DAG structure, $G_{scm}$.
    \item Define each node as an affine transformation of its parent nodes followed by a randomly sampled  activation function.
    \item Assign each node a noise distribution \(p(\epsilon)\), sampled from a meta-distribution over noise distributions.
\end{enumerate}
This defines a structured causal model, where each node is defined as $
z_i = f_i(z_{\mathrm{pa}(i)}, \epsilon_i).
$
Given the sampled SCM, a tabular dataset is generated as follows:

\begin{enumerate}
    \item Sample root variables (causes) from a mixture of Gaussian, multinomial, and Zipfian distributions.
    \item Propagate samples through the DAG with additive noise, i.e., \( \tilde{x} = x + \epsilon_i \), where \( \epsilon_i \sim p(\epsilon) \) and $i$ is the index of the node.
    \item Randomly select $d_{in}$ nodes as input features \( x_i \) and one node as the continuous target \( \hat{y} \).
    \item
    With probability 0.5 generate mixed-type datasets, where each feature is independently converted to a discrete variable with probability \( p \sim \mathcal{U}(0,1) \). Otherwise, we generate homogeneous datasets: with probability 0.5 all features are discretized, and with probability 0.5 all features remain continuous. For categorical features, the number of categories is sampled as
    \( k \sim \max(\text{round}(\text{Gamma}(1, 8)), 1) \), and values are obtained mapping values to interval indices, following the procedure used in TabPFN.
    \item Discretize the continuous target \( \hat{y} \) into a binary label \( y \) either by thresholding at its median or by assigning samples in the lower and upper quantiles (e.g., 25th and 75th percentiles) with probability 0.5 for each case.
\end{enumerate}

Finally, this procedure allows the generation of synthetic tabular datasets while also enabling sampling from two distributions $p(x \mid y=0)$ and $p(x \mid y=1)$. This forms the basis for the generation of normal samples and SCM anomalies.

\subsection{Perturbation-based Anomalies}

In addition to SCM-based labels, we generate anomalies through feature-level perturbations. This approach produces anomalous samples independently of the SCM label and complements the SCM anomaly. For instance, SCM anomalies rarely represent large feature-wise deviations. These are large feature spikes are captured by the perturbation-based anomalies. 

\begin{enumerate}
    \item Generate a dataset using the SCM prior (steps 1–4 above).
    \item Sample support and query sets according to the supervision regime.
    \item Z-score normalize the features using statistics from the support set.
    \item Apply corruption to half of the query samples labeled as anomalous:
    
    \begin{enumerate}
        \item \emph{Categorical corruption}  
        Sample a corruption rate \( \lambda \sim \mathcal{U}(0, \lambda_{\max}) \) where $\lambda_{\max}=1$. For each categorical feature, replace its value with a uniformly sampled category with probability \( \lambda \).

        \item \emph{Continuous corruption}  
        For each sample, draw a sparsity rate \( s \sim \mathcal{U}(0,1) \). For each feature \( j \), sample a mask \( m_j \sim \text{Bernoulli}(s) \). Apply the Gaussian noise:
        \[
        \tilde{x}_j = x_j + m_j\sigma_j \epsilon_j, \quad \epsilon_j \sim \mathcal{N}(0,1),
        \]
        where \( \sigma_j \sim \text{LogUniform}(\sigma_{\min}, \sigma_{\max}) \).
    \end{enumerate}
\end{enumerate}

\subsection{Synthetic Tasks Hyperparameters}

In this section, we summarize the hyperparameters in task generation. 
The SCM hyperparameters are listed in Tables~\ref{tab:scm_cat_hps} and~\ref{tab:scm_cont_hps}, 
while hyperparameters for general task construction and perturbation-based anomalies are provided in Table~\ref{tab:tasks_pert_hps}.

\begin{table}[!h]
\centering
\caption{Categorical Hyperparameters and configuration settings for SCMs. Each is sampled independently with uniform probability per dataset.}
\begin{tabular}{ll}
\toprule
Hyperparameter & Choices \\
\midrule
Share node-wise noise & True, False \\
Apply blockwise dropout & True, False \\
Preserve feature order & True, False \\
Blockwise feature selection & True, False \\
Activation function & Tanh, ELU, Sigmoid, Identity, Sine, Softplus, Heaviside \\
\bottomrule
\end{tabular}
\label{tab:scm_cat_hps}
\end{table}

\begin{table}[!h]
\centering
\caption{Continuous hyperparameter distributions used in synthetic SCM generation. TNLU(\( \mu, \tilde{\mu}, \text{round}, \text{min}\)) denotes log-uniform distribution over parameter \(\mu\) followed by a truncated log normal distribution with minimal values and potentially rounding the sampled value to the nearest integer.}
\begin{tabular}{llllll}
\toprule
Hyperparameter & Distribution & & & & \\
\midrule
MLP weight dropout & \multicolumn{4}{l}{0.9 \( \text{Beta}(a, b) \), where \( a, b \sim \mathcal{U}(0.1,5.0) \)} \\
\midrule
 & & Max \(\mu\) & Min \(\tilde{\mu}\) & Round & Min \\
\midrule
MLP depth & TNLU & 8 & 1 & Yes & 2 \\
MLP width (hidden nodes) & TNLU & 180 & 5 & Yes & 4\\
Number of Causes (Nodes at layer 1) & TNLU & 12 & 1 & Yes & 1\\
SCM Node noise Std. & TNLU & 0.3 & 0.0001 & No & 0.0\\
SCM Weight Std. & TNLU & 10.0 & 0.01 & No & 0.0\\
\bottomrule
\end{tabular}
\label{tab:scm_cont_hps}
\end{table}

\begin{table}[!tph]
\caption{Hyperparameter distributions or constants used in task generation and perturbation-based anomalies.}
\centering
\setlength{\tabcolsep}{10pt}
\begin{tabular}{ll}
\toprule
Hyperparameter & Distribution / Constant \\
\midrule
Dataset dimension $d_{in}$ & \(\mathcal{U}(2, 512)\) \\
Support set size $N_{s}$ & \(\mathcal{U}(5, 12000)\) \\
Query set size $N_{q}$ & 1024 \\
Perturbation Noise Std. (\(\sigma_j\)) & $\mathrm{LogUniform}(0.1, 10.0)$ \\
Categorical corruption rate $\lambda$ & \(\mathcal{U}(0,1)\) \\
\bottomrule
\end{tabular}
\label{tab:tasks_pert_hps}
\end{table}

\newpage

\subsection{Pseudocode for Tasks Construction across Supervision Regimes}

Algorithm~\ref{alg:scenarios} provides a high-level description of the task construction procedure across supervision regimes.

\begin{algorithm}[!h]
\caption{Synthetic task construction across supervision regimes}
\label{alg:scenarios}
\begin{algorithmic}[1]
\Require support size $N_s$, query size $N_q$, supervision regime weights $\pi$
\Ensure support set $(X_s, Y_s)$ and query set $(X_q, Y_q)$

\State Sample scenario $r \sim \mathrm{Categorical}(\pi)$, 
where $\pi = (\pi_{\text{one-class}}, \pi_{\text{unsup}}, \pi_{\text{semi}})$
\State Sample contamination ratio $\rho \sim \mathcal{U}(0, 0.4)$
\State Generate dataset $\mathcal{D}$ using the SCM prior

\Statex \textbf{Support set construction}
\If{$r = \text{one-class}$}
    \State Sample $N_s$ normal samples from $\mathcal{D}$
    \State Set $Y_s \gets 0$
\ElsIf{$r = \text{unsupervised}$}
    \State Sample $N_s$ samples with contamination ratio $\rho$
    \State Mark \(N_a = \text{round}(\rho N_s)\) anomalous samples. 
    \State Set $Y_s \gets -1$
\ElsIf{$r = \text{semi-supervised}$}
    \State Sample $N_s$ samples with contamination ratio $\rho$
    \State Mark \(N_a = \text{round}(\rho N_s)\) anomalous samples. 
    \State Sample supervision ratio $\rho_{\text{sup}} \sim \mathcal{U}(0,1)$
    \State Mark  \(N_{sup} = \text{round}(\rho_{sup} N_a)\) anomalous samples to reveal labels. 
    \State Set $Y_s = 1$ for revealed anomalies and $Y_s = -1$ otherwise
\EndIf

\Statex \textbf{Query set construction}
\State Sample $N_q/2$ normal samples and $N_q/2$ anomalous samples
\State Assign anomaly types for anomalous samples.
\State Replace half of query anomalous samples with SCM anomalies.
\State Replace the other half with perturbation-based anomalies.

\Statex \textbf{Post-processing}
\State Z-score normalize all samples using statistics from the support set
\State Apply perturbation-based corruption where required

\State \Return $(X_s, Y_s), (X_q, Y_q)$
\end{algorithmic}
\end{algorithm}

Particularly, in line 2, the contamination ratio $\rho$ ranges from little to no anomaly contamination in the near 0 case to the extreme case of 40\% contamination. Line 14 to 16 shows supervision levels ranging from 0\% anomalies labeled to full partial supervision where 100\% of anomalies are labeled. Here, we also see that the unsupervised scenario is a special case of semi-supervised scenario where no supervision is available and one-class setting is a special case of unsupervised setting where there is no contamination. This gradual transition from one setting to another motivates the need for a model that can adapt seamlessly to the full spectrum of supervision and contamination levels.

\newpage

\section{Additional Visualizations of SCM and Perturbation-based Anomalies}

\begin{figure}[!h]
    \centering
    \begin{subfigure}{0.495\textwidth}
        \centering
        \includegraphics[width=\textwidth]{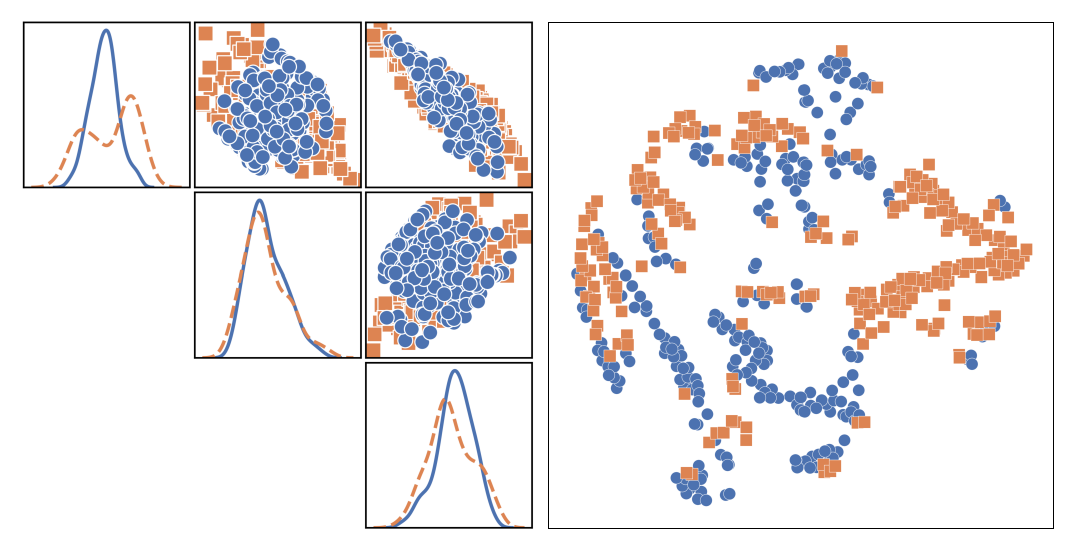}
    \end{subfigure}
    \begin{subfigure}{0.495\textwidth}
        \centering
        \includegraphics[width=\textwidth]{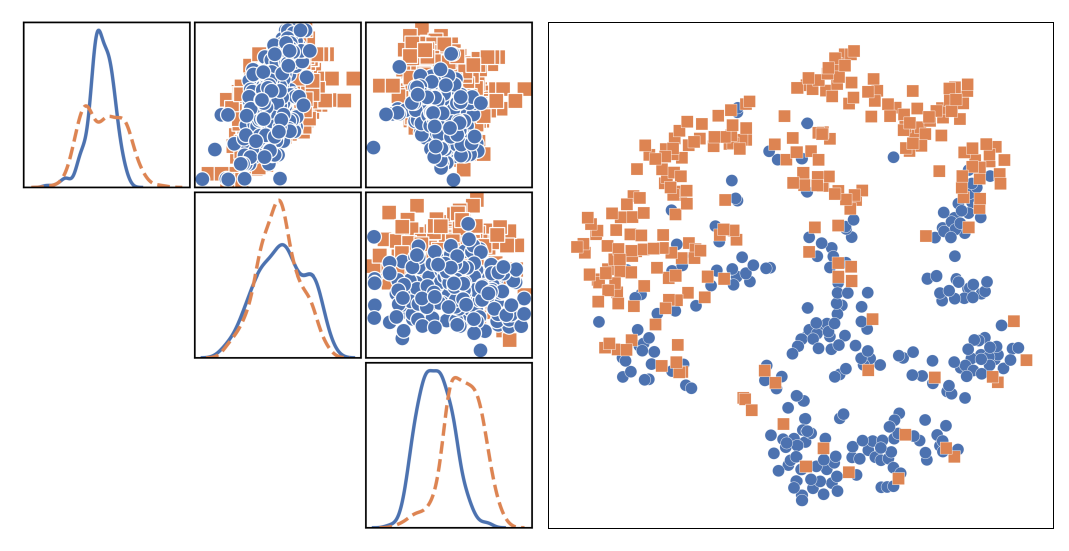}
    \end{subfigure}
    \begin{subfigure}{0.495\textwidth}
    \centering
    \includegraphics[width=\textwidth]{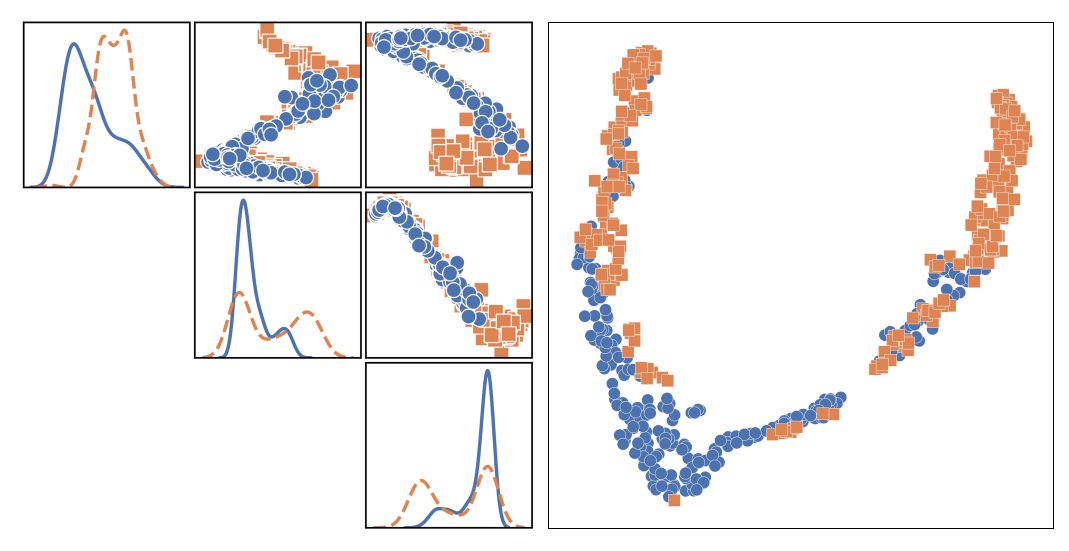}
    \end{subfigure}
    \begin{subfigure}{0.495\textwidth}
    \centering \includegraphics[width=\textwidth]{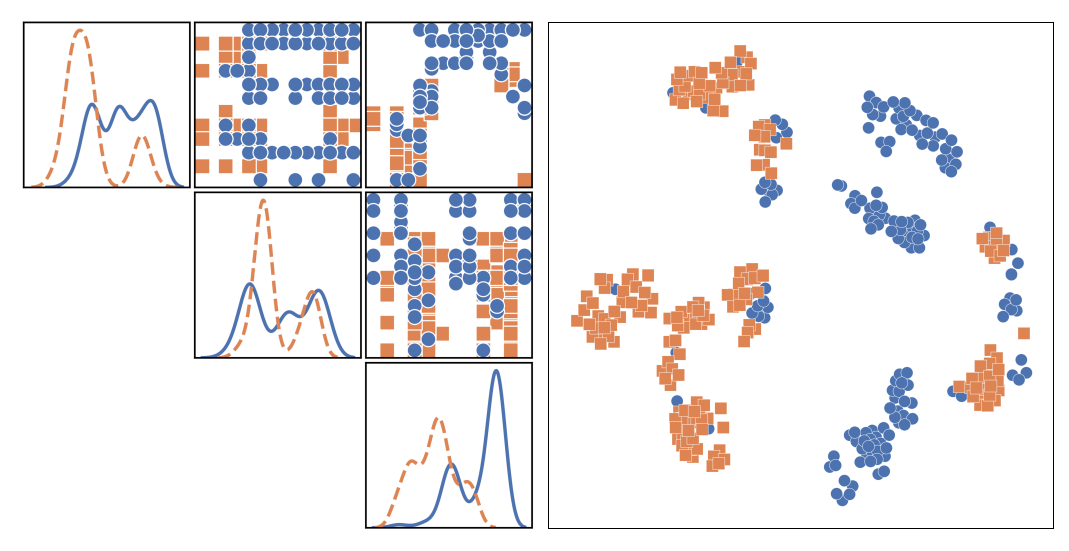}
    \end{subfigure}
    \caption{Feature interaction and t-SNE plots of normal samples and SCM anomalies.}
    \label{fig:scm_anom_visualization2}
\end{figure}

\begin{figure}[!h]
    \centering
    \begin{subfigure}{0.495\textwidth}
        \centering
        \includegraphics[width=\textwidth]{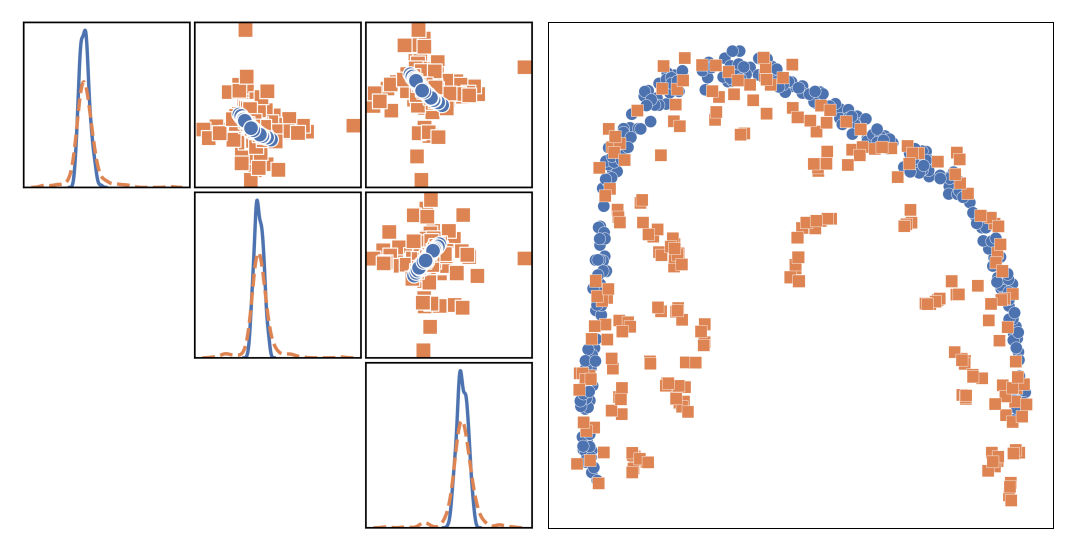}
    \end{subfigure}
    \begin{subfigure}{0.495\textwidth}
        \centering
        \includegraphics[width=\textwidth]{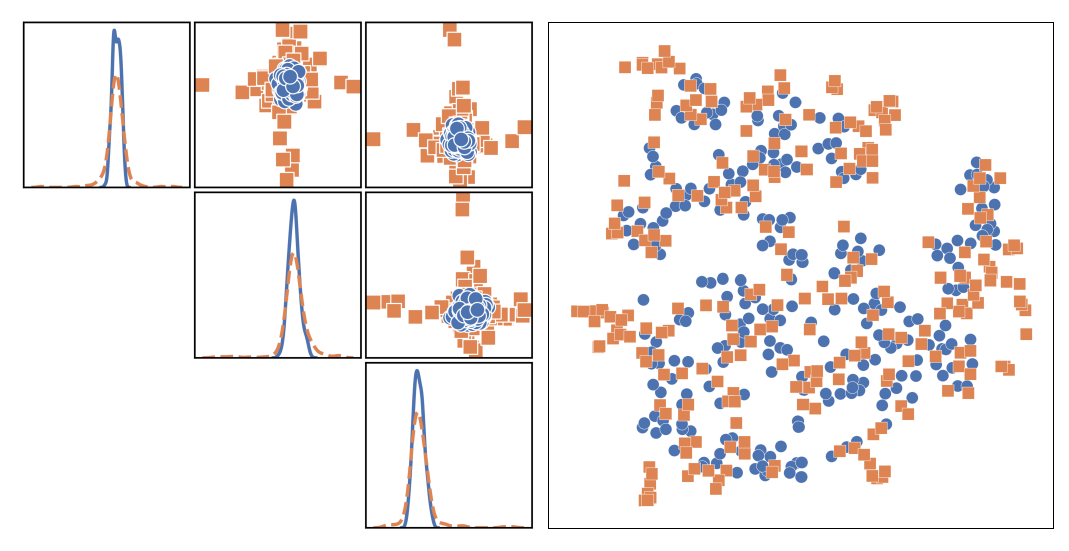}
    \end{subfigure}
    \begin{subfigure}{0.495\textwidth}
    \centering
    \includegraphics[width=\textwidth]{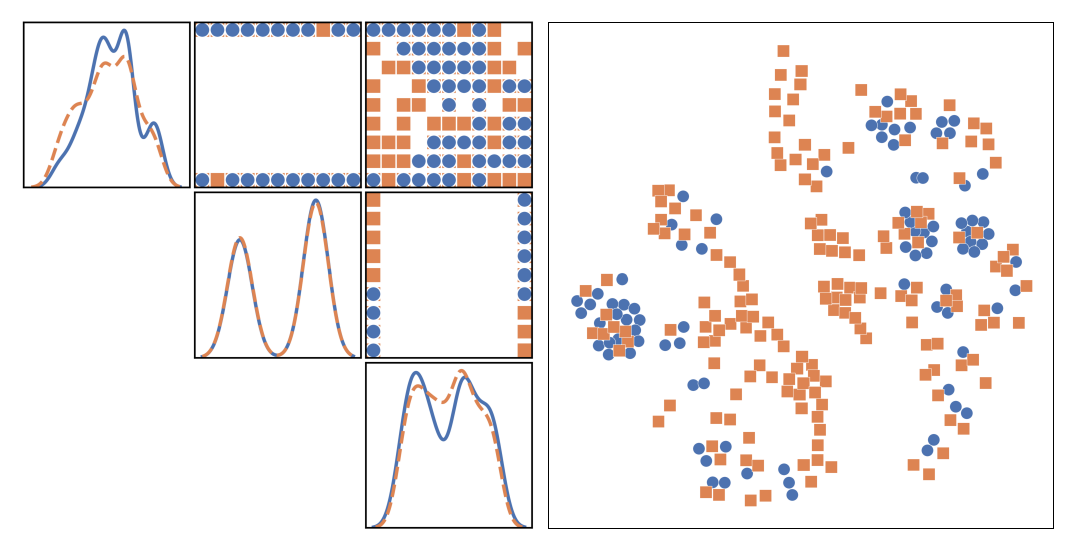}
    \end{subfigure}
    \begin{subfigure}{0.495\textwidth}
    \centering \includegraphics[width=\textwidth]{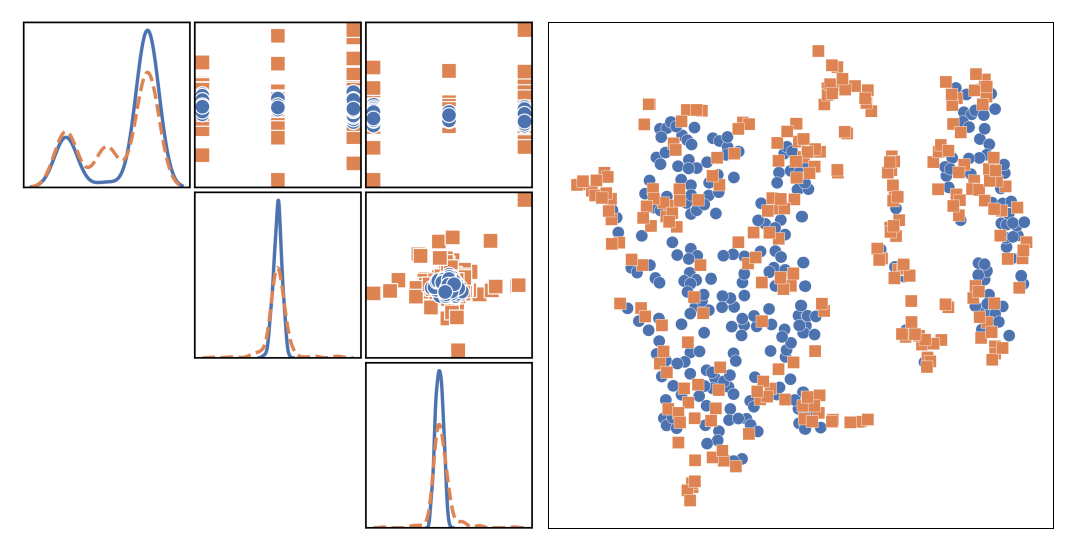}
    \end{subfigure}
    \caption{Feature interaction and t-SNE plots of normal and perturbation-based anomaly samples. The orange rectangles are anomalies and the blue circles represent normal samples.}
    \label{fig:pert_anom_visualization2}
\end{figure}

\newpage

\section{Model and Synthetic Tasks Validation}

We validate our models and synthetic tasks using a small development set and do not perform extensive per-dataset tuning. This design choice reflects our goal of evaluating ICLAD as a general-purpose in-context learner, rather than optimizing performance for individual datasets.

\subsection{Development Datasets}
We validate ICLAD on a development set consisting of 7 datasets that are not part of the ADBench test suite. This separation helps prevent meta-overfitting to the test benchmarks during development. The development set includes 5 real-world datasets from \cite{shenkarAnomalyDetectionTabular2021a}, namely \emph{abalone}, \emph{arrhythmia}, \emph{ecoli}, \emph{mulcross}, and \emph{seismic}. In addition, we construct two synthetic two-dimensional datasets based on the Scikit-learn \emph{moons} and \emph{circles} generators.

\subsubsection{Supervision Regimes}
For each dataset, we construct three evaluation scenarios corresponding to our supervision regimes. These scenarios are designed to match the testing protocol we used for ADBench.

\emph{One-class setting} Following the one-class protocols from previous works, we split the normal class into two disjoint subsets. One subset is used as the training set that contains only normal samples, while the other subset is combined with all anomaly samples to form the test set.

\emph{Unsupervised setting} We construct the training set by sampling with replacement from the full dataset, resulting in a dataset with anomaly contamination. The test set is the full dataset.

\emph{Semi-supervised setting} Starting from the unsupervised setting, we randomly label 10\% of the anomalies in the training set, while the remaining samples remain unlabeled.

\clearpage

\section{Additional Experimental Results}

\subsection{One-class Setting Box Plots}

\begin{figure}[!htbp]
    \centering

    \begin{subfigure}{0.95\textwidth}
        \centering
        \includegraphics[width=\textwidth]{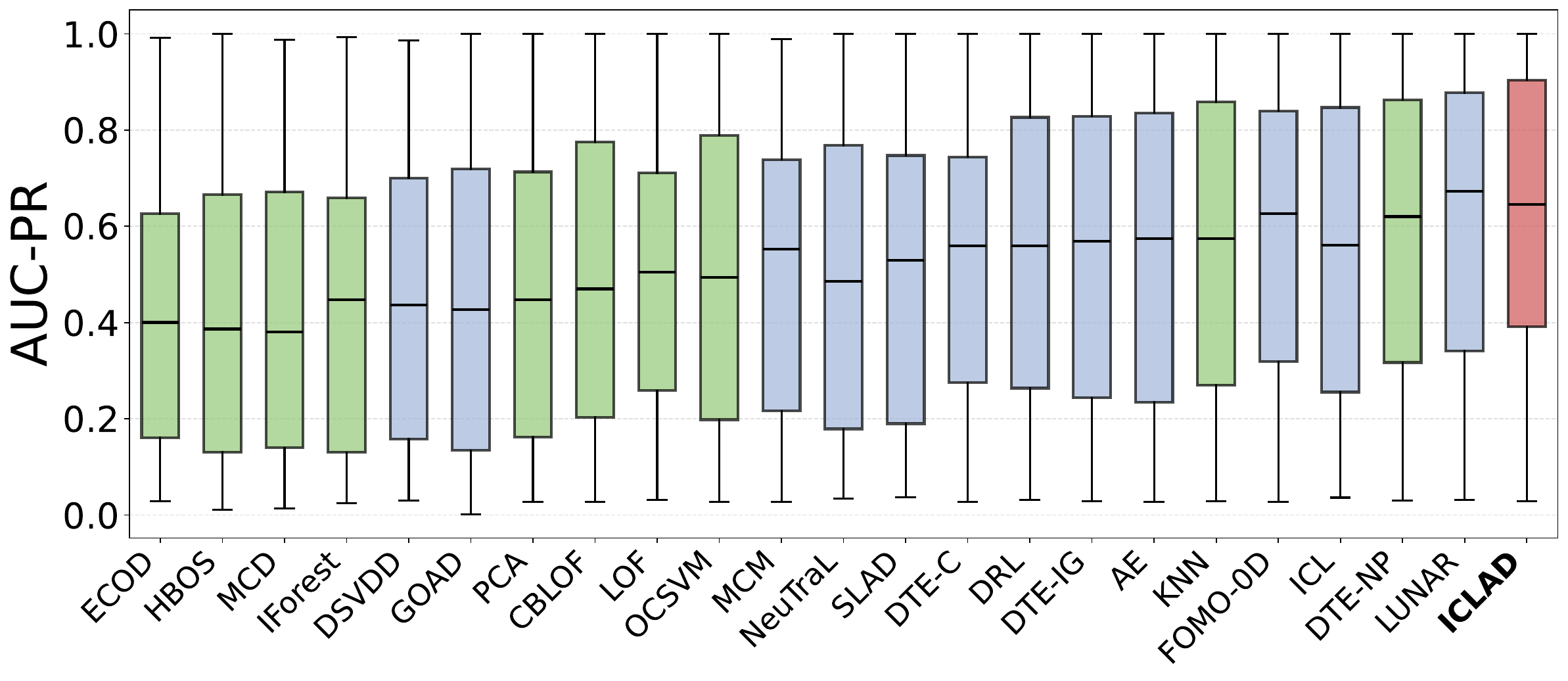}
    \end{subfigure}

    \begin{subfigure}{0.95\textwidth}
        \centering
        \includegraphics[width=\textwidth]{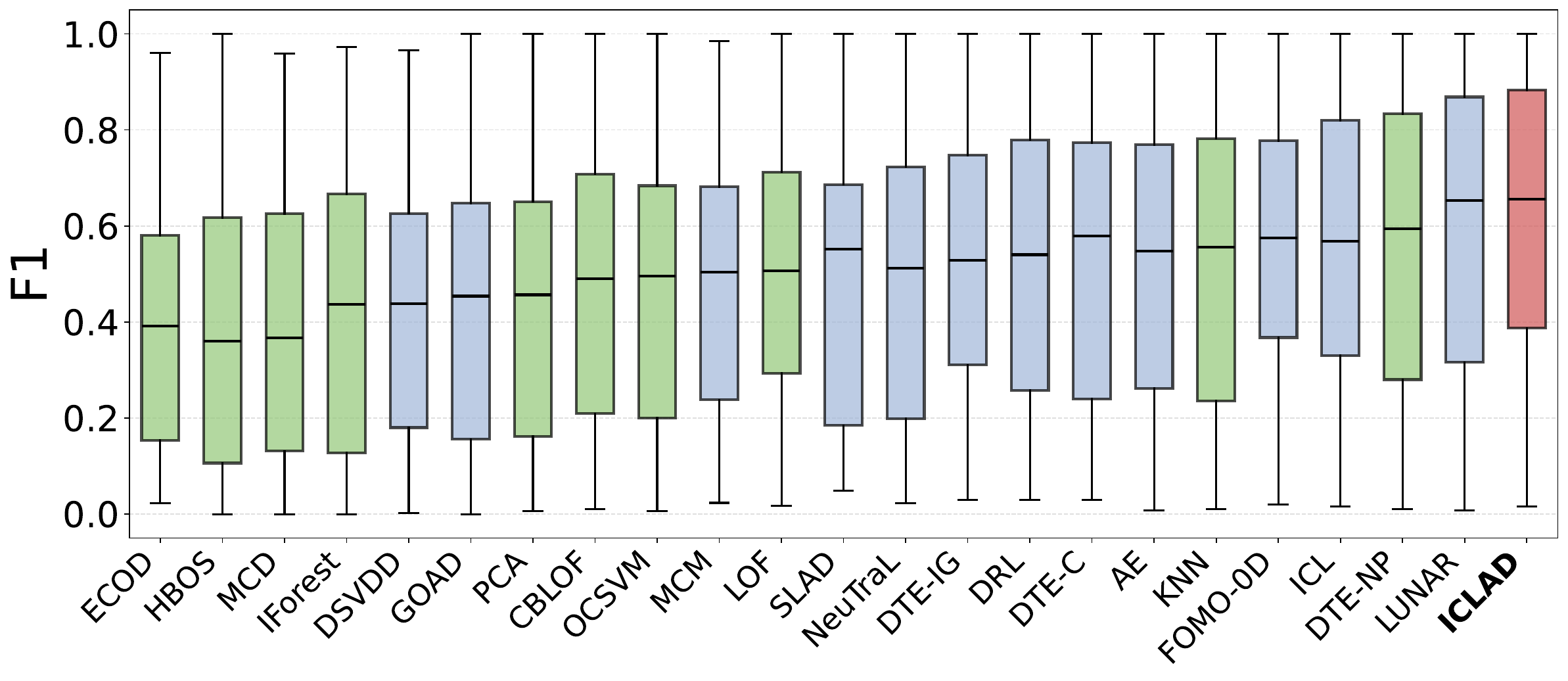}
    \end{subfigure}

    \caption{Boxplots for one-class setting. Boxes show the interquartile range (IQR) with medians indicated by the center line and whiskers extending to 1.5 times IQR. Models are ordered by average AUC-PR or F1 and color-coded by family: classical (green), deep learning (blue), and ICLAD (red).}
    \label{fig:iclad_oc_boxplots}
\end{figure}

Figures~\ref{fig:iclad_oc_boxplots} shows the boxplots for AUC-PR and F1 in the one-class setting. The overall trends for both metrics are consistent with the AUC-ROC results reported in the main paper. We highlight the noticeable improvement in F1 over the baselines methods.

\newpage

\subsection{Unsupervised Setting Box Plots}

\begin{figure}[!htbp]
    \centering

    \begin{subfigure}{0.95\textwidth}
        \centering
        \includegraphics[width=\textwidth]{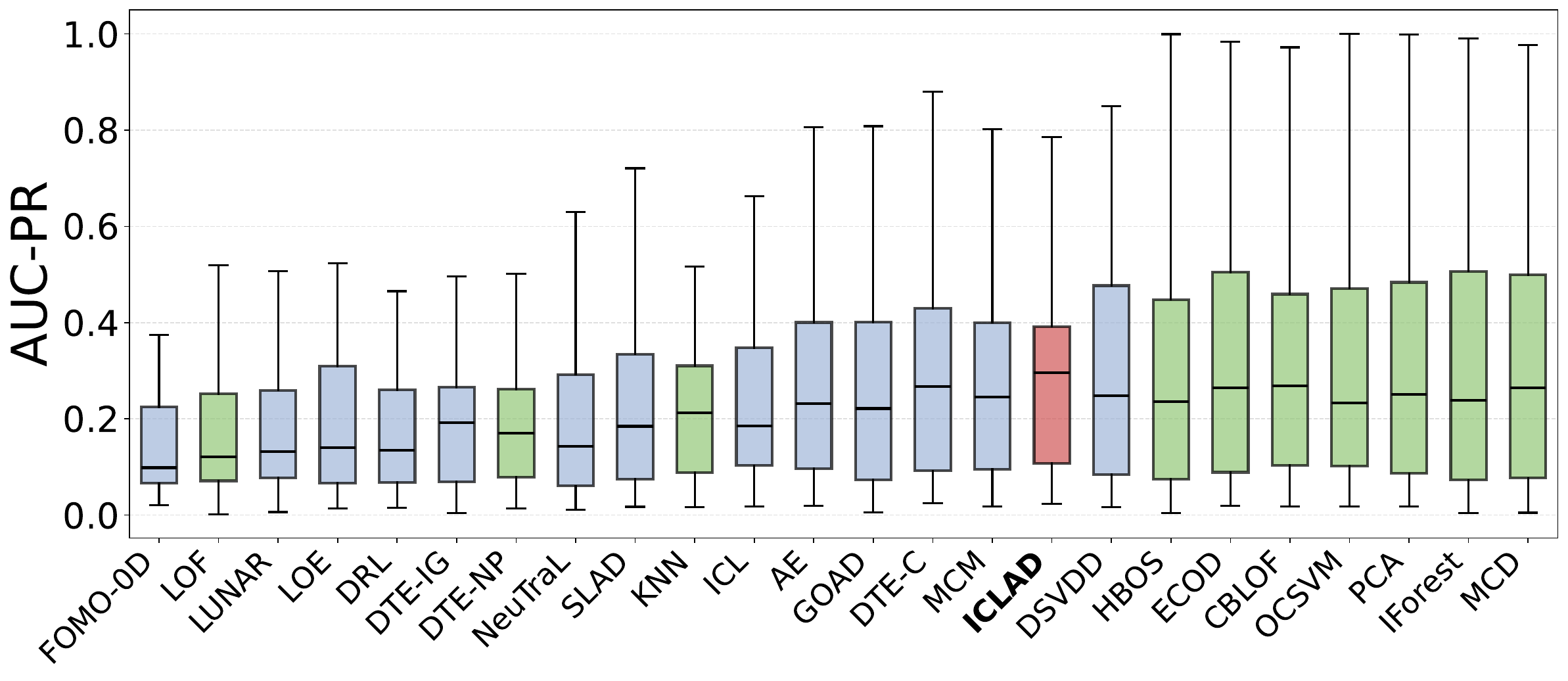}
    \end{subfigure}

    \begin{subfigure}{0.95\textwidth}
        \centering
        \includegraphics[width=\textwidth]{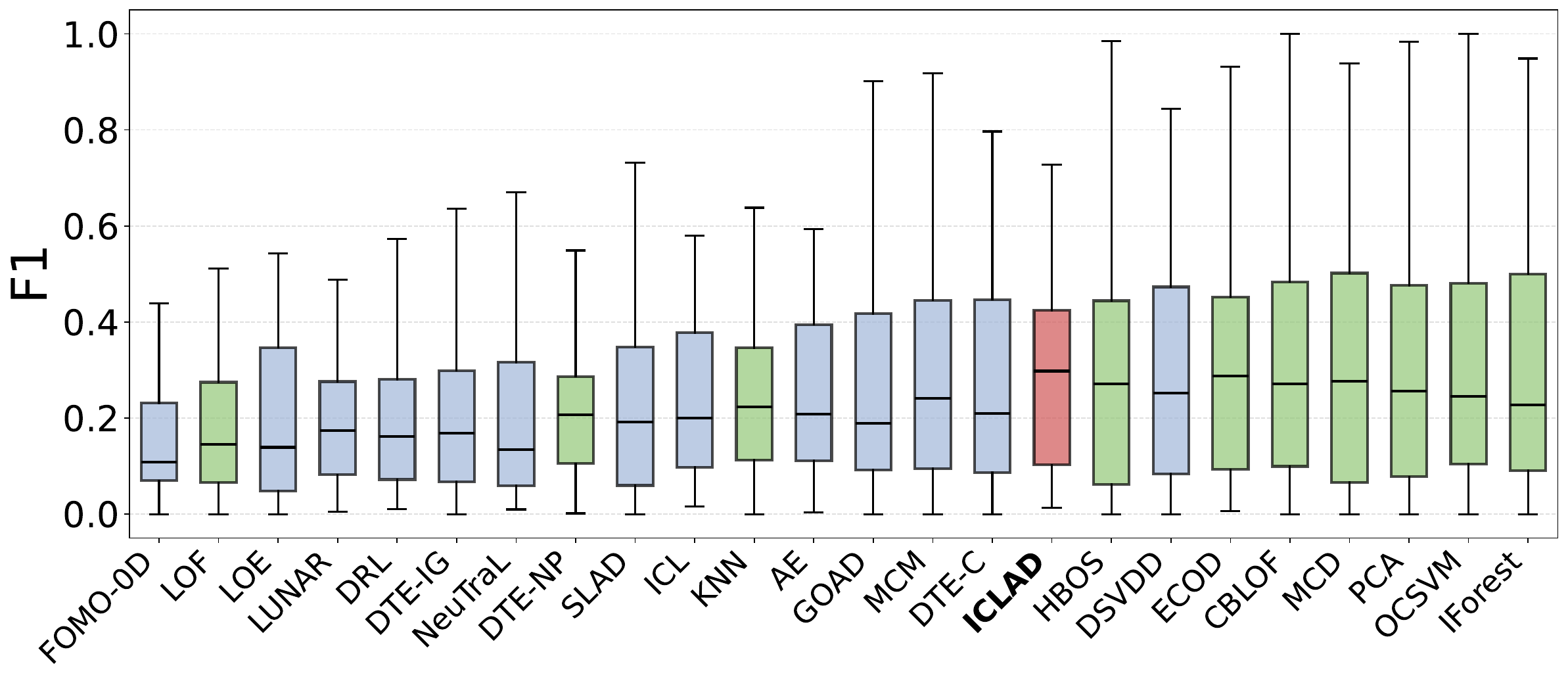}
    \end{subfigure}

    \caption{Boxplots for the unsupervised setting. Boxes show the interquartile range (IQR) with medians indicated by the center line and whiskers extending to 1.5 times IQR. Models are ordered by average AUC-PR or F1 and color-coded by family: classical (green), deep learning (blue), and ICLAD (red).}
    \label{fig:iclad_unsup_boxplots}
\end{figure}

Figure~\ref{fig:iclad_unsup_boxplots} suggests that the AUC-PR and F1 results for ICLAD are on par with the best performing deep learning approaches but fall short of strong classical baselines. In contrast with the strong AUC-ROC results, this discrepancy suggest that contamination still degrades the model’s ability to form a sharp decision boundary, affecting precision in the high anomaly score region. This result shows the persistent challenge in the unsupervised regime for deep learning approaches. Nevertheless, our framework partially bridges this gap by maintaining strong global ranking performance under contamination. This challenge also underlines the importance of supervision for deep learning approaches where even minimal labeled anomalies consistently lead to substantial improvements.

\newpage

\subsection{One-class Setting Critical Difference Diagrams}

\begin{figure}[!htbp]
    \centering

    \begin{subfigure}[b]{\textwidth}
        \centering
        \includegraphics[width=\linewidth]{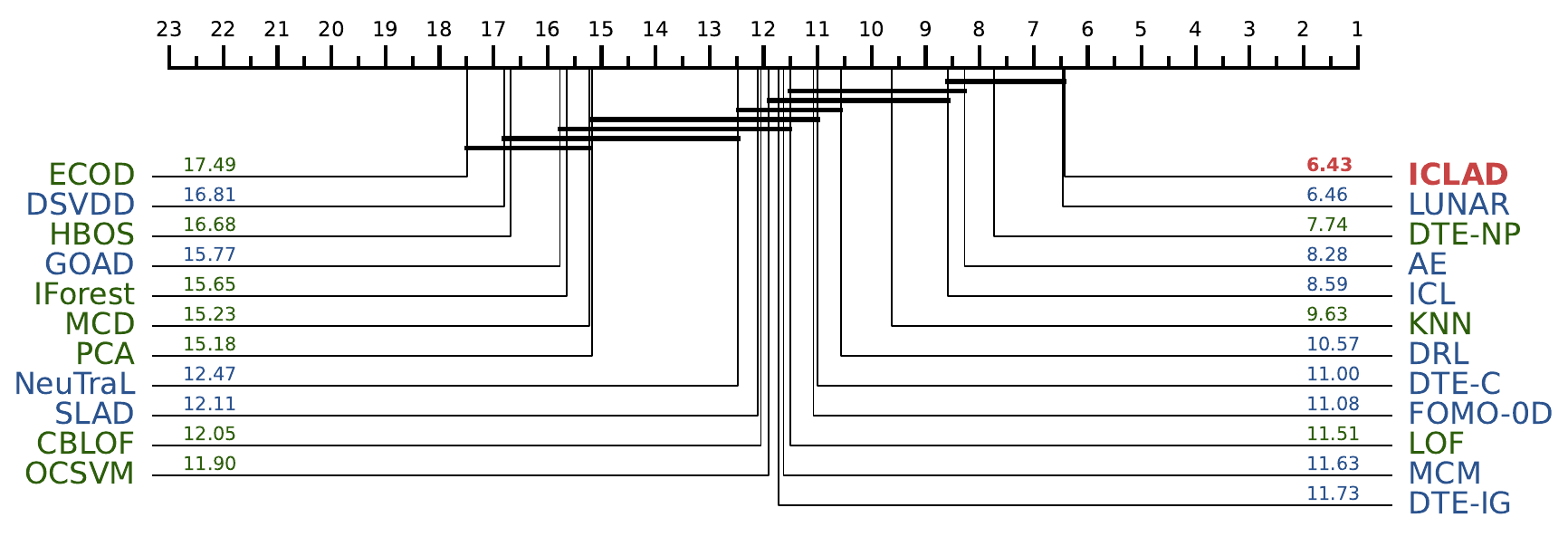}
        \caption{AUC-PR Ranks}
        \label{fig:iclad_oc_cdd_aucpr}
    \end{subfigure}

    \begin{subfigure}[b]{\textwidth}
        \centering
        \includegraphics[width=\linewidth]{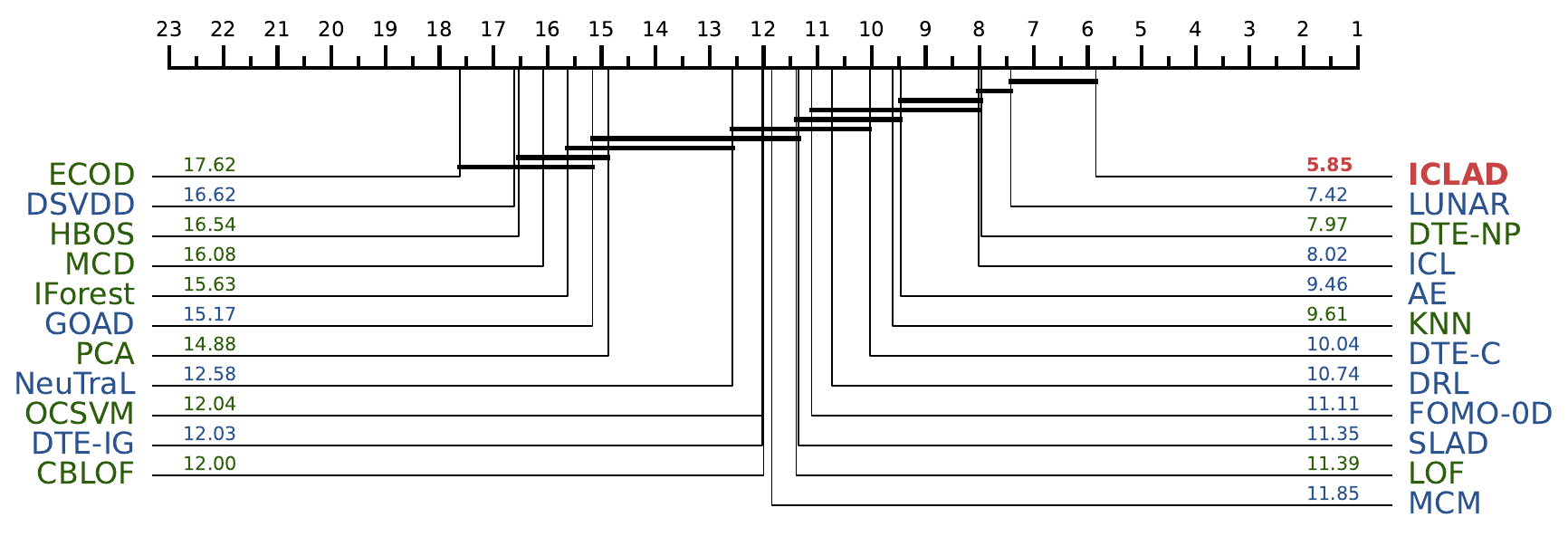}
        \caption{F1 Ranks}
        \label{fig:iclad_oc_cdd_f1}
    \end{subfigure}

    \caption[ICLAD One-Class Setting Critical Difference Diagrams]{
        Critical difference diagrams for the one-class setting.
    }
    \label{fig:iclad_oc_cdds}
\end{figure}

Figure~\ref{fig:iclad_oc_cdds} further strengthens ICLAD superior performance in the one-class regime, particularly in terms of F1. While ICLAD is statistically competitive with the top-performing method LUNAR, it achieves a lower (better) average rank and improves upon strong baselines such as kNN and DTE-NP.

\newpage

\subsection{Unsupervised Setting Critical Difference Diagrams}

\begin{figure}[!htbp]
    \centering

    \begin{subfigure}[b]{\textwidth}
        \centering
        \includegraphics[width=\linewidth]{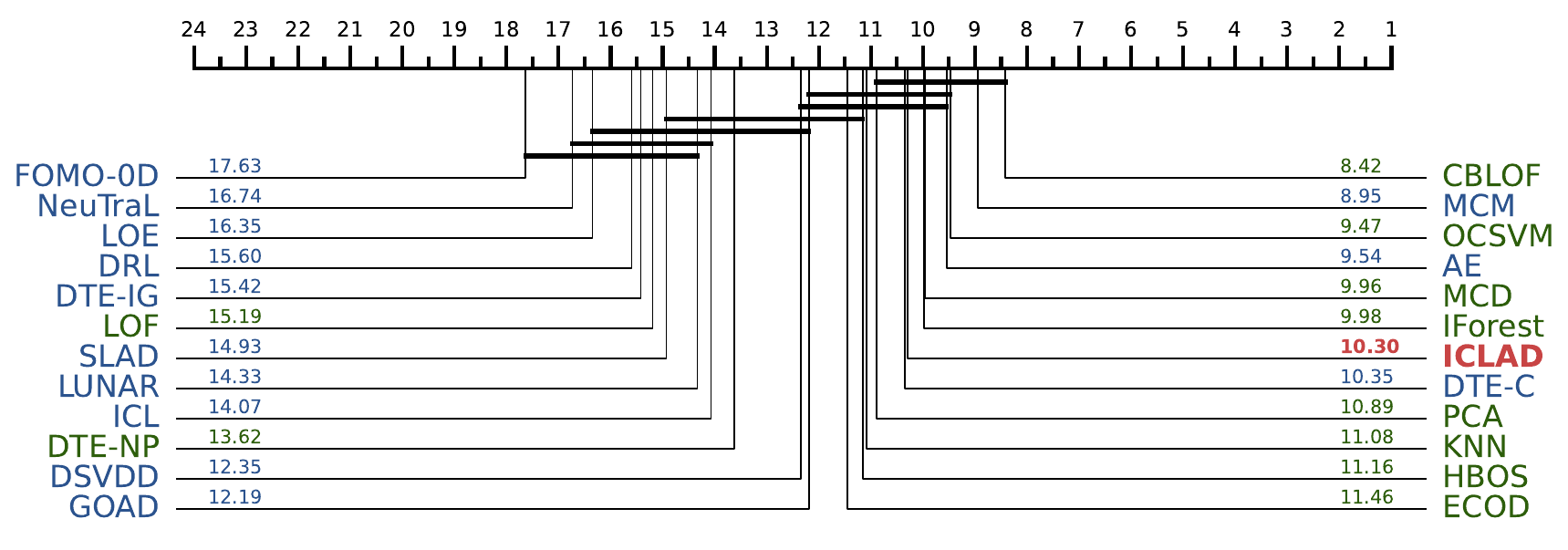}
        \caption{AUC-PR Ranks}
        \label{fig:iclad_unsup_cdd_aucpr}
    \end{subfigure}

    \begin{subfigure}[b]{\textwidth}
        \centering
        \includegraphics[width=\linewidth]{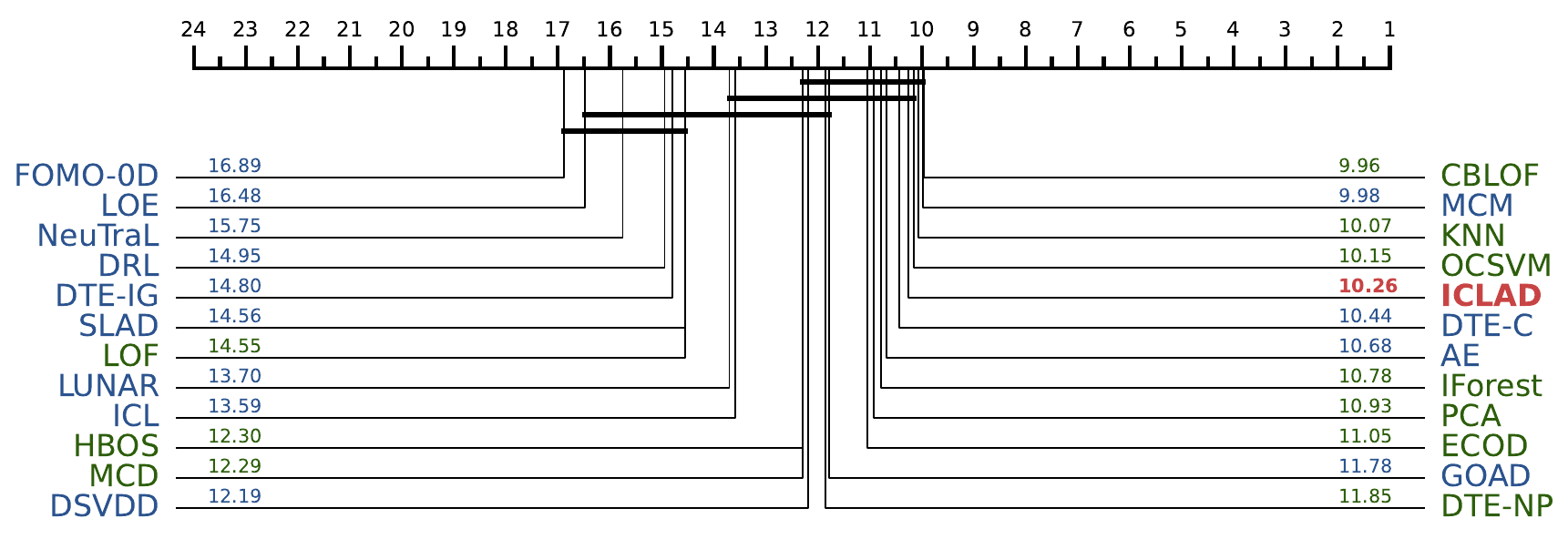}
        \caption{F1 Ranks}
        \label{fig:iclad_unsup_cdd_f1}
    \end{subfigure}

    \caption[ICLAD One-Class Setting Critical Difference Diagrams]{
        Critical difference diagrams for the unsupervised setting.
    }
    \label{fig:iclad_unsup_cdd}
\end{figure}

The critical difference diagram in Figure~\ref{fig:iclad_unsup_cdd} shows comparable performance across all methods in terms of AUC-PR and F1, with CBLOF achieving the best overall performance. ICLAD lies within the same statistical group as the top-performing methods, indicating competitive performance. Consistent with the bar plot results, deep learning models generally exhibit a loss in precision under anomaly contamination, whereas ICLAD demonstrates comparatively stronger robustness to contamination.

\newpage

\subsection{Semi-supervised Setting F1 Curves}

\begin{figure}[h]
    \centering
    \includegraphics[width=0.8\linewidth]{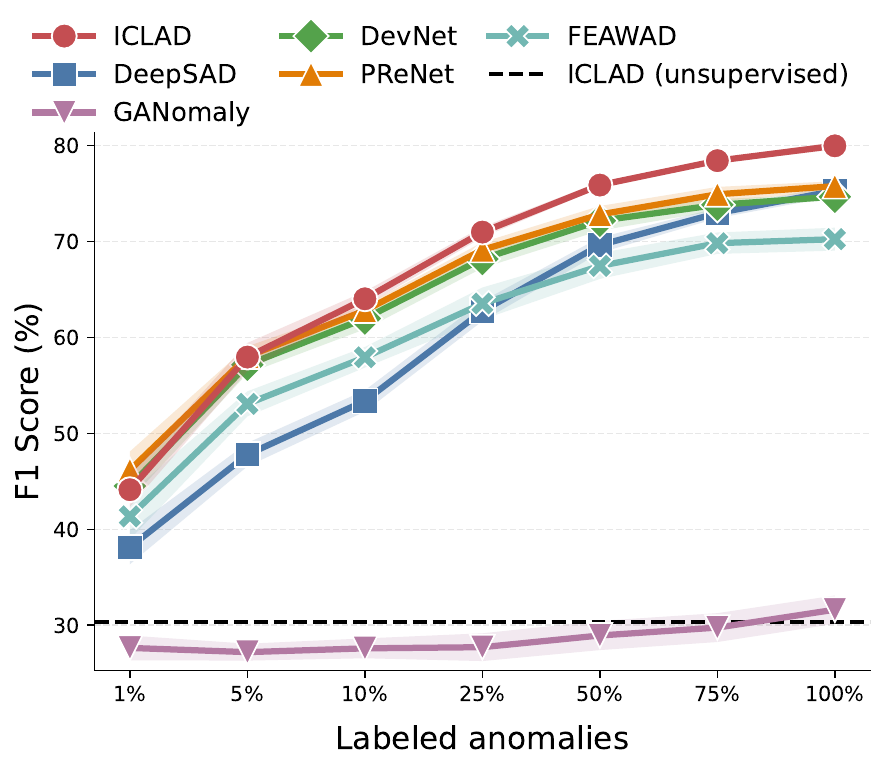}

    \caption[Effect of Labeled Anomaly Availability on ICLAD in the Semi-supervised Setting]{
        Semi-supervised performance curves measured by F1 (\%) against ratio of labeled anomalies.
    }
    \label{fig:iclad_semi_f1}
\end{figure}

Figure~\ref{fig:iclad_semi_f1} show the improvements over F1 with increasing ratio of labeled anomalies provided to the semi-supervised models. As with AUC-ROC and AUC-PR, all methods except for GANomaly benefit significantly from partial supervision. Although ICLAD's F1 performance is generally comparable to PreNet and DevNet when 1\% of anomalies are labeled, ICLAD's performance gap with stronger baselines start to manifest when 10\% of anomalies are available with the gap increasing with more labeled anomalies. This trend shows that ICLAD utilizes labeled more effectively than other semi-supervised models.    

\newpage

\subsection{Total Time vs Average AUC-ROC plots}

\begin{figure}[!h]
    \centering
    \includegraphics[width=0.75\textwidth]{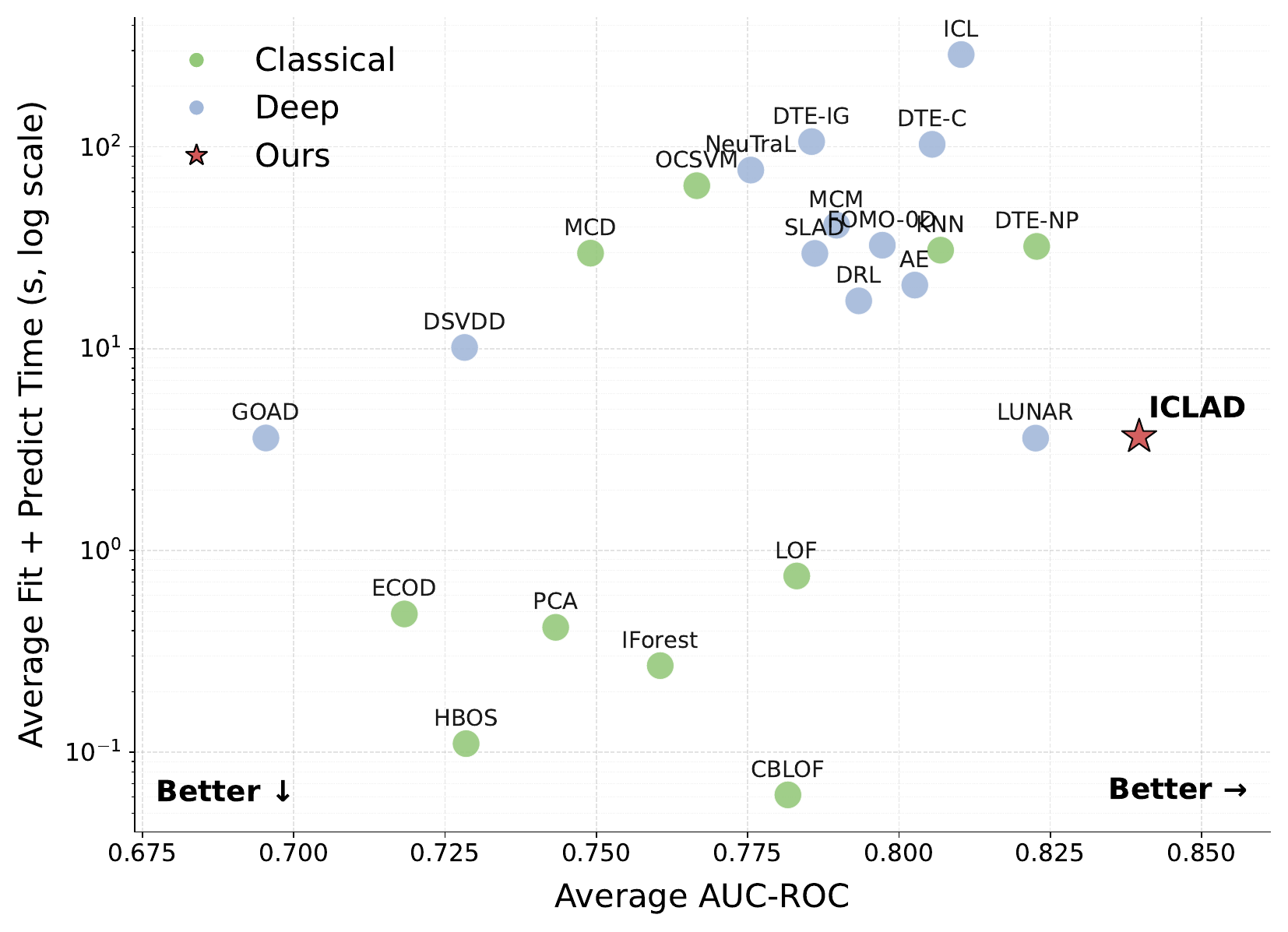}

    \caption[Effect of Labeled Anomaly Availability on ICLAD in the Semi-supervised Setting]{
        Average Total Time (Fit + Predict) in seconds vs. Average AUC-ROC in one-class setting. 
    }
    \label{fig:iclad_oc_time_vs_aucroc}
\end{figure}

\begin{figure}[!h]
    \centering
    \includegraphics[width=0.75\textwidth]{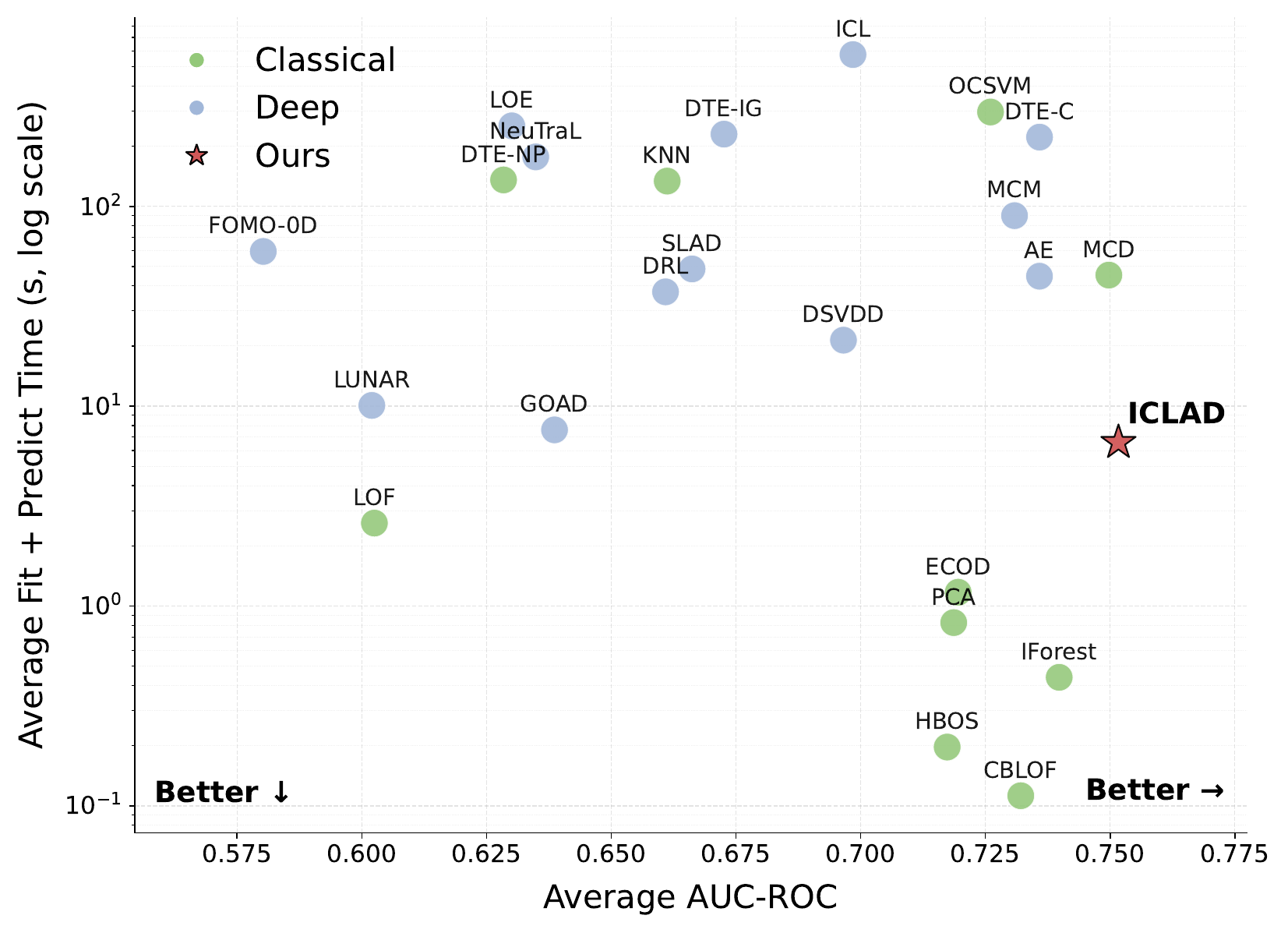}

    \caption[Effect of Labeled Anomaly Availability on ICLAD in the Semi-supervised Setting]{
        Average Total Time (Fit + Predict) in seconds vs. Average AUC-ROC in the unsupervised setting. 
    }
    \label{fig:iclad_unsup_time_vs_aucroc}
\end{figure}

\section{Benchmark Dataset Details}

\begin{table}[!hpt]
\centering
\setlength{\tabcolsep}{3.5pt}

\begin{tabular}{lrrrrrl}
\toprule
Dataset & \#Samples & \#Feat. & \#Anom. & \%Anom. & Category \\
\midrule
ALOI & 49534 & 27 & 1508 & 3.04 & Image \\
annthyroid & 7200 & 6 & 534 & 7.42 & Healthcare \\
backdoor & 95329 & 196 & 2329 & 2.44 & Network \\
breastw & 683 & 9 & 239 & 34.99 & Healthcare \\
campaign & 41188 & 62 & 4640 & 11.27 & Finance \\
cardio & 1831 & 21 & 176 & 9.61 & Healthcare \\
Cardiotocography & 2114 & 21 & 466 & 22.04 & Healthcare \\
celeba & 202599 & 39 & 4547 & 2.24 & Image \\
census & 299285 & 500 & 18568 & 6.20 & Sociology \\
cover & 286048 & 10 & 2747 & 0.96 & Botany \\
donors & 619326 & 10 & 36710 & 5.93 & Sociology \\
fault & 1941 & 27 & 673 & 34.67 & Physical \\
fraud & 284807 & 29 & 492 & 0.17 & Finance \\
glass & 214 & 7 & 9 & 4.21 & Forensic \\
Hepatitis & 80 & 19 & 13 & 16.25 & Healthcare \\
http & 567498 & 3 & 2211 & 0.39 & Web \\
InternetAds & 1966 & 1555 & 368 & 18.72 & Image \\
Ionosphere & 351 & 33 & 126 & 35.90 & Oryctognosy \\
landsat & 6435 & 36 & 1333 & 20.71 & Astronautics \\
letter & 1600 & 32 & 100 & 6.25 & Image \\
Lymphography & 148 & 18 & 6 & 4.05 & Healthcare \\
magic.gamma & 19020 & 10 & 6688 & 35.16 & Physical \\
mammography & 11183 & 6 & 260 & 2.32 & Healthcare \\
mnist & 7603 & 100 & 700 & 9.21 & Image \\
musk & 3062 & 166 & 97 & 3.17 & Chemistry \\
optdigits & 5216 & 64 & 150 & 2.88 & Image \\
PageBlocks & 5393 & 10 & 510 & 9.46 & Document \\
pendigits & 6870 & 16 & 156 & 2.27 & Image \\
Pima & 768 & 8 & 268 & 34.90 & Healthcare \\
satellite & 6435 & 36 & 2036 & 31.64 & Astronautics \\
satimage-2 & 5803 & 36 & 71 & 1.22 & Astronautics \\
shuttle & 49097 & 9 & 3511 & 7.15 & Astronautics \\
skin & 245057 & 3 & 50859 & 20.75 & Image \\
smtp & 95156 & 3 & 30 & 0.03 & Web \\
SpamBase & 4207 & 57 & 1679 & 39.91 & Document \\
speech & 3686 & 400 & 61 & 1.65 & Linguistics \\
Stamps & 340 & 9 & 31 & 9.12 & Document \\
thyroid & 3772 & 6 & 93 & 2.47 & Healthcare \\
vertebral & 240 & 6 & 30 & 12.50 & Biology \\
vowels & 1456 & 12 & 50 & 3.43 & Linguistics \\
Waveform & 3443 & 21 & 100 & 2.90 & Physics \\
\bottomrule
\end{tabular}

\caption{Characteristics of classical tabular datasets from ADBench (Part 1).}
\label{tab:adbench_datasets_tabular1}
\end{table}

\begin{table}[!hpt]
\centering
\setlength{\tabcolsep}{8pt}

\begin{tabular}{lrrrrrl}
\toprule
Dataset & \#Samples & \#Feat. & \#Anom. & \%Anom. & Category \\
\midrule
WBC & 223 & 9 & 10 & 4.48 & Healthcare \\
WDBC & 367 & 30 & 10 & 2.72 & Healthcare \\
Wilt & 4819 & 5 & 257 & 5.33 & Botany \\
wine & 129 & 13 & 10 & 7.75 & Chemistry \\
WPBC & 198 & 33 & 47 & 23.74 & Healthcare \\
yeast & 1484 & 8 & 507 & 34.16 & Biology \\
\bottomrule
\end{tabular}

\caption{Characteristics of classical tabular datasets from ADBench (Part 2).}
\label{tab:adbench_datasets_tabular2}
\end{table}

\begin{table}[!h]
\centering
\setlength{\tabcolsep}{6.5pt}

\begin{tabular}{lrrrrrl}
\toprule
Dataset & \#Samples & \#Feat. & \#Anom. & \%Anom. & Category \\
\midrule
CIFAR10 & 5263 & 512 & 263 & 5.00 & Image \\
FashionMNIST & 6315 & 512 & 315 & 5.00 & Image \\
MNIST-C & 10000 & 512 & 500 & 5.00 & Image \\
MVTec-AD & 5354 & 512 & 1258 & 23.50 & Image \\
SVHN & 5208 & 512 & 260 & 5.00 & Image \\
\midrule
Agnews & 10000 & 768 & 500 & 5.00 & NLP \\
Amazon & 10000 & 768 & 500 & 5.00 & NLP \\
Imdb & 10000 & 768 & 500 & 5.00 & NLP \\
Yelp & 10000 & 768 & 500 & 5.00 & NLP \\
20newsgroups & 11905 & 768 & 591 & 4.96 & NLP \\
\bottomrule
\end{tabular}

\caption{Characteristics of embedding-based datasets (image and text) from ADBench.}
\label{tab:adbench_datasets_embeddings}
\end{table}

\end{document}